\documentclass[twoside]{article}

\usepackage[T1]{fontenc}

\usepackage[accepted]{aistats2024}
\usepackage[hyphens]{url}
\usepackage{pdfpages}
\usepackage{xcolor}
\usepackage{bold-extra}
\usepackage{graphicx}
\usepackage{amssymb}

\usepackage{placeins}
\usepackage[round]{natbib}

\begin{document}



\twocolumn[
\runningtitle{How does GPT-2 Predict Acronyms? Understanding a Circuit via Mechanistic Interpretability}
\aistatstitle{How does GPT-2 Predict Acronyms? Extracting and Understanding a Circuit via Mechanistic Interpretability}

\aistatsauthor{ Jorge García-Carrasco \And Alejandro Maté \And  Juan Trujillo }

\aistatsaddress{ Lucentia Research, Department of Software and Computing Systems, University of Alicante } ]

\begin{abstract}
    Transformer-based language models are treated as black-boxes because of their large number of parameters and complex internal interactions, which is a serious safety concern. Mechanistic Interpretability (MI) intends to reverse-engineer neural network behaviors in terms of human-understandable components. In this work, we focus on understanding how GPT-2 Small performs the task of predicting three-letter acronyms. Previous works in the MI field have focused so far on tasks that predict a single token. To the best of our knowledge, this is the first work that tries to mechanistically understand a behavior involving the prediction of multiple consecutive tokens. We discover that the prediction is performed by a circuit composed of 8 attention heads ($\sim 5\%$ of the total heads) which we classified in three groups according to their role. We also demonstrate that these heads concentrate the acronym prediction functionality. In addition, we mechanistically interpret the most relevant heads of the circuit and find out that they use positional information which is propagated via the causal mask mechanism. We expect this work to lay the foundation for understanding more complex behaviors involving multiple-token predictions. 

    % The popularity of Transformer-based language models is rising exponentially due to its impressive performance on a wide range of tasks. However, they are treated as black-boxes because of their large number of parameters and complex internal interactions, which is a serious safety concern. Mechanistic Interpretability (MI) intends to reverse-engineer neural network behaviors in terms of human-understandable components. In this work, we focus on understanding how GPT-2 Small performs the task of predicting three-letter acronyms. Previous works in the MI field have focused so far on tasks that predict a single token. To the best of our knowledge, this is the first work that tries to mechanistically understand a behavior involving the prediction of multiple consecutive tokens. We discover that the prediction is performed by a circuit composed of 8 attention heads ($\sim 5\%$ of the total heads) which we classified in three groups according to their role. \todo{Ablating every other head does preserve the performance} We also demonstrate that these heads concentrate the acronym prediction functionality. In addition, we mechanistically interpret the most relevant heads of the circuit and find out that they use positional information which is propagated via the causal mask mechanism. We expect this work to lay the foundation for understanding more complex behaviors involving multiple-token predictions. 
\end{abstract}

\section{INTRODUCTION}

Scaling up the size of Language Models based on the Transformer architecture \citep{vaswani2017attention, brown2020language} has been shown to greatly improve its performance on a wide range of tasks. Because of this, the use of Large Language Models (LLMs) on high-impact fields such as in medicine \citep{thirunavukarasu2023large, zhang2023continuous} is increasingly growing and is expected to keep growing even larger. However, these models are treated as black-boxes due to the fact that they have a large number of parameters and complex internal interactions, hampering our ability to understand its behavior. This is a considerable concern with regards to the safety of Artificial Intelligence (AI) systems, as using a model without knowing its internal heuristics or algorithms can derive in unexpected outcomes such as privacy issues \citep{li2023multi} or harmful behavior \citep{wei2023jailbroken} among others. 

Mechanistic Interpretability (MI) aims to tackle this issue by interpreting behaviors in terms of human-understandable algorithms or concepts \citep{elhage2021mathematical, olsson2022context, elhage2022toy, nanda2023progress}. In other words, it tries to reverse-engineer the large amount of parameters that compose the model into understandable components, which essentially increases the trustworthiness of the model. MI has been successfully applied to explain different tasks on transformer-based models. For example, \cite{wang2022interpretability} discover the circuit responsible for the Indirect Object Identification task (IOI) in GPT-2 Small, composed by 26 attention heads grouped in 7 different classes. Similarly, \cite{hanna2023does} use MI techniques to explain how GPT-2 Small performs the greater-than operation on a single task and test if the discovered circuit generalizes to other contexts. Likewise, \cite{docstring} discovered how a smaller 4-layer transformer model predicted argument names on a docstring. 

In summary, works like the ones mentioned above are proof that mechanistic analysis can be used to shine light into the inner workings of language models. As MI is a young field, the current focus is on developing methods and understanding behaviors on relatively smaller models to lay a solid foundation that will be used to interpret increasingly larger models. In fact, preliminary studies have already appeared discussing whether current MI techniques are scalable to larger models, with optimistic results \citep{lieberum2023does}.

In this work, we contribute to the growing body of MI works by focusing on understanding how GPT-2 Small performs the task of predicting three-letter acronyms (e.g. \texttt{"The Chief Executive Officer"} $\rightarrow$ \texttt{"CEO"}). We have mainly chosen this task because it consists on predicting three consecutive tokens, in contrast to previous existing work which focused on single-token prediction. To the best of our knowledge, this is the first work that applies MI to understand a behavior involving the prediction of multiple tokens. Hence, we expect that the work presented here serves as a starting point for understanding more complex behaviors that involve predicting multiple tokens.

More specifically, we will adopt a \emph{circuits} perspective \citep{elhage2021mathematical, olah2020zoom} and identify the components of the model that are responsible for the behavior under study via a series of systematic \emph{activation patching} \citep{meng2022locating} experiments. Our contributions can be summarized as follows:
\begin{itemize}
    \item We discover the circuit responsible for three-letter acronym prediction on GPT-2 Small. The circuit is composed by 8 attention heads ($\sim 5\%$ of GPT-2's heads) which we classified on three groups according to their role.
    \item We evaluate the circuit by ablating the rest of components of the model and show that the performance is preserved and even slightly improved when isolating the discovered 8-head circuit.
    \item We interpret the main components of the circuit, which we term \emph{letter mover heads} by reverse-engineering their parameters.
    \item We also found that \emph{letter mover heads} make use of positional information, mainly derived from the attention probabilities due to the causal mask mechanism instead of the positional embeddings.
\end{itemize}

The remainder of this paper is structured as follows. In Section \ref{sec:background}, the required background and the problem statement are presented. Section \ref{sec:circuit} describes the procedure used to discover the circuit responsible for three-letter acronym prediction as well as the role of each component, followed by an evaluation of the circuit. Section \ref{sec:LMH} delves into mechanistically interpreting \emph{letter mover heads} as well as studying how these heads use positional information. Finally, the conclusions about the work are presented in Section \ref{sec:conclusions}. 

% Our work first begins by clearly defining the task at hand and a performance metric. Then, we curate a three-letter acronyms dataset that will be used to perform the following experiments. Then, we discover the circuit responsible for this task. We do so via activation patching techniques, carefully corrupting different components to discover their role on the complete circuit. Once that the circuit is discovered, we perform ablation experiments to isolate the discovered circuit, and show that in some situations, the performance can even be improved. Finally, we delve into the learned parameters of the model to understand how they work.

\section{BACKGROUND}\label{sec:background}

In this section we briefly present the transformer and the used notation, the task of study and how to evaluate the performance of the model on that task.

\subsection{Model and Notation}

GPT-2 Small \citep{radford2019language} is a 117M parameter decoder-only transformer architecture composed by 12 transformer blocks containing 12 attention heads followed by an MLP, each component preceded by Layer Normalization \citep{ba2016layer}. The input to the model is a sequence of $N$ consecutive tokens which are embedded into $x_0 \in \mathbb{R}^{N \times d}$ via a learned embedding matrix $W_E \in \mathbb{R}^{V \times d}$, where $V$ is the size of the vocabulary. Similarly, positional embeddings are added to $x_0$.

Following the notation presented in \cite{elhage2021mathematical}, $x_0$ is the initial value of the \emph{residual stream}, where all the components of the model read from and write to. Specifically, if $h_{ij}$ denotes the $j$th attention head at layer $i$, the $i$th attention layer will update the residual stream as $x_{i+1} = x_i + \sum_j h_{ij}(x_i)$ (omitting layer normalization). Each attention head is parameterized by the matrices $W_Q^{ij}, W_K^{ij}, W_V^{ij} \in \mathbb{R}^{d \times d/H}$ and $W_O^{ij} \in \mathbb{R}^{d/H \times d}$, where $H$ is the number of heads in a single layer, which can be arranged into the QK and OV matrices $W_{QK}^{ij} = W_Q^{ij} W_K^{ij}$, $W_{OV}^{ij} = W_V^{ij} W_O^{ij}$. The QK matrix contains information about which tokens the head attends to, whereas the OV matrix is related to what the head writes into the residual stream.

Finally, the resulting vector is unembedded via a unembedding matrix, which in the case of GPT-2 is tied to the embedding matrix (i.e. $W_U = W_E^T$) to obtain a vector $y \in \mathbb{R}^{N \times V}$ where $y_{ij}$ represents the logits of the $j$th token of the vocabulary for the prediction following the $i$th token of the sequence.

\subsection{Task Description}

% We choose the prediction of three-letter acronyms as our task of study, e.g. we enter a prompt such as \texttt{"The Cane Knee Lender ("} and the model has to predict the acronym \texttt{"CKL"}.

We will focus on the task of predicting three-letter acronyms. To evaluate whether GPT-2 is able to properly perform this task or not, we curated a dataset of 800 acronyms. It is important to remark that this dataset will \textbf{not} be used to re-train the model, but to perform experiments and identify the underlying circuit associated to the task of study. In other words, our aim is to detect a circuit responsible for a concrete task on an LLM that has already been trained in a general, self-supervised way. Hence, in order to isolate the behavior of study and reduce the amount of noise, we made each acronym to meet the following characteristics:
\begin{itemize}
    \item Each word must be composed by two tokens, the first being only composed by the capital letter and its preceding space (e.g. \texttt{"| C|ane|"})
    \item The acronym must be tokenized by exactly three tokens, each for one letter of the acronym (e.g. \texttt{"|C|K|L|"})
\end{itemize}

% POSSIBLE APPENDIX IF PAPER IS TOO LARGE

In order to build the dataset, we took a public list of the most frequently-used common nouns in English \citep{noun_list} containing a total of $6775$ nouns. However, building the dataset is not as easy as choosing three random words and tokenizing them according to our imposed characteristics: words have to be tokenized as GPT-2 naturally expects to stay in-distribution. GPT-2 uses byte-pair encoding (BPE) tokenization \citep{sennrich2015neural}, a technique that tokenizes according to the most frequent substring. This means that common substrings/words such as \verb|"ABC"| or \verb|" Name"| are encoded as a single token, hence reducing the amount of possible nouns and acronyms to use on our dataset. Taking this into account, the building procedure was the following:

\begin{enumerate}
    \item \textbf{Nouns:} We took the list of 6775 nouns and filtered out the words that did not meet the characteristics (i.e. each word of the acronym must be composed by two tokens, the first being only composed by the capital letter and its preceding space), leaving us with 381 nouns.
    \item \textbf{Acronyms:} We tokenized the $P^R(26,3) = 26^3 = 17576$ possible 3-letter acronyms and checked which were naturally tokenized as three separate tokens, which reduced the amount of possible acronyms to 2740. As common nouns beginning with the letter \texttt{X}, \texttt{Q} or \texttt{U} are rare, we also excluded acronyms containing that letter, resulting in a total of 1154 possible combinations.
    \item \textbf{Dataset:} Finally, we built the dataset by (i) sampling one of the 152 possible acronyms (e.g. \texttt{WVZ}) and randomly sampling three of the 381 nouns, one for every letter of the acronym (e.g. \texttt{Wreck}, \texttt{Vibe} and \texttt{Zipper}). Notice that we can build much more than 800 samples in this way, but we chose this size because of computational constraints. As a reference, \cite{hanna2023does} curated a dataset of 490 datapoints to properly identify a circuit.
\end{enumerate}

In summary, this results in a dataset composed by prompts with the structure:
\begin{center}
    \texttt{"|The|C1|T1|C2|T2|C3|T3| (|A1|A2|A3|"}
\end{center}
where \texttt{Ci} is the token encoding the capital letter of the $i$th word (together with its preceding space), \texttt{Ti} is the remainder of the word, and \texttt{Ai} is the $i$th letter of the acronym. Therefore, the task consists on predicting \texttt{A1}, \texttt{A2} and \texttt{A3} given the previous context.

The reason for choosing a list of nouns was because (i) it is the common way to build acronyms and (ii) the type of word (noun, adjective, etc.) does not affect the result obtained. We can confirm these results since we have also experimented with synthetic words by taking a random token \texttt{"| A|"} where \texttt{A} can be any capital letter, followed by one to three random tokens containing just lowercase letters and found the same results that will be presented in this work.

Also, one important concern is that the model could have memorized popular acronyms (e.g. \texttt{The Central Processing Unit (CPU)}). However, the acronyms built with our procedure are rare (e.g. \texttt{The Wreck Vibe Zipper (WVZ)}). We took this decision to ensure that the model has not been trained with these samples, implying that the discovered circuit generalizes to samples outside of the training dataset and does not just memorize common acronyms.

% \begin{enumerate}
%     \item We tokenized the 3276 possible combinations of 3-letter acronyms and checked which were naturally tokenized as three separate tokens, which reduced the amount of possible acronyms to 407. As common nouns beginning with the letter \texttt{X} are rare, we also excluded acronyms containing that letter, resulting in a total of 152 possible combinations.
%     \item Similarly, we tokenized every noun of the list, discarding the ones that did not meet the required characteristics. This reduced the list to a total of 381 nouns.
%     \item Now, we can easily build a suitable sentence by randomly choosing one of the possible acronyms, and three random nouns from the list that begin with the corresponding letters.
% \end{enumerate}

% This results in a dataset composed by prompts with the structure:
% \begin{center}
%     \texttt{"|The|C1|T1|C2|T2|C3|T3| (|A1|A2|A3|"}
% \end{center}
% where \texttt{Ci} is the token encoding the capital letter of the $i$th word (together with its preceding space), \texttt{Ti} is the remainder of the word, and \texttt{Ai} is the $i$th letter of the acronym. Therefore, the task consists on predicting \texttt{A1}, \texttt{A2} and \texttt{A3} given the previous context.

\subsection{Evaluation}

In order to quantitatively evaluate the ability of GPT-2 on the task under study, we will compute the \emph{logit difference} between the correct letter and the incorrect letter with the highest logits, for each of the three letters of the acronym:

\begin{equation}
    \text{logit\_diff}_i = \text{logits}_{a_i} - \underset{l \in \mathcal{L} \backslash \{a_i\}}{\text{max}} \text{logits}_l
\end{equation}

where $a_i$ is the correct prediction for the $i$th letter of the acronym and $\mathcal{L}$ is the set of possible predictions, which in the case of acronym prediction, is the set of capital letters. GPT-2 has an average logit difference across every letter and sentence of the dataset of $2.22$, which translates to an average $\sim 90.2 \%$ probability difference. Overall, this result provides quantitative evidence supporting that GPT-2 is indeed able to perform three-letter acronym prediction.

\section{A CIRCUIT FOR 3-LETTER ACRONYM PREDICTION}\label{sec:circuit}

Now that the task has been clearly defined and checked that GPT-2 is indeed able to perform it, we will discover the circuit responsible for this behavior, evaluate it and understand the components that compose such circuit. The following experiments were performed by using both \emph{PyTorch} \citep{paszke2019pytorch} and \emph{TransformerLens} \citep{nandatransformerlens2022} with a 40GB A100 GPU. A repository containing the code required to reproduce the experiments and figures can be found in \url{https://github.com/jgcarrasco/acronyms_paper}.

\subsection{Discovering the circuit}

In order to discover which components form the circuit responsible for three-letter acronym prediction, we will perform a systematic series of \emph{activation patching} experiments, first presented in \cite{meng2022locating}. The idea of activation patching is to patch (i.e. replace) the activations of a given component with the activations obtained by running the model on a \emph{corrupted prompt}. If the metric degrades when patching a component, it means that it is relevant to the task under study, therefore enabling us to locate the circuit.

In this case, we have carefully performed activation patching with three different types of corruption. For each of the $i$th letter prediction, we (i) randomly resample the $i$th word, (ii) randomly resample the words previous to the $i$th word and (iii) randomly resample the acronym letters previous to the $i$th letter. This will allow us to better track the flow of information and the role of each component. Table \ref{tab:corruption-examples} shows the different types of corruption for the prediction of the third letter on a sample prompt. We will perform the corresponding activation patching experiments for each of the three letters on parallel.
% In the experiments, we will simultaneously perform activation patching experiments for each of the three letters by properly corrupting the corresponding tokens. 

%\todo{Cambiar tipografia Ci?}

\begin{table}[htbp]
\caption{Example of prompt corruption for the third letter prediction $i=3$ (this is also performed for $i=1, 2$)} \label{tab:corruption-examples}
\begin{center}
\begin{tabular}{ll}
\textbf{Corruption Type}  &\textbf{Prompt} \\
\hline \\
Original&\texttt{The Cane Knee Lender (CK} \\
Current Word& \texttt{The Cane Knee \textbf{Tandem} (CK} \\
Previous Words&\texttt{The \textbf{Ego Icy} Lender (CK} \\
Previous Letters&\texttt{The Cane Knee Lender (\textbf{BJ}}\\
All corruptions&\texttt{The \textbf{Ego Icy Tandem} (\textbf{BJ}}\\
\end{tabular}
\end{center}
\end{table}

\subsubsection{Corrupting the Current Word}

\begin{figure}[htbp]
    \centering
    \includegraphics[width=\linewidth]{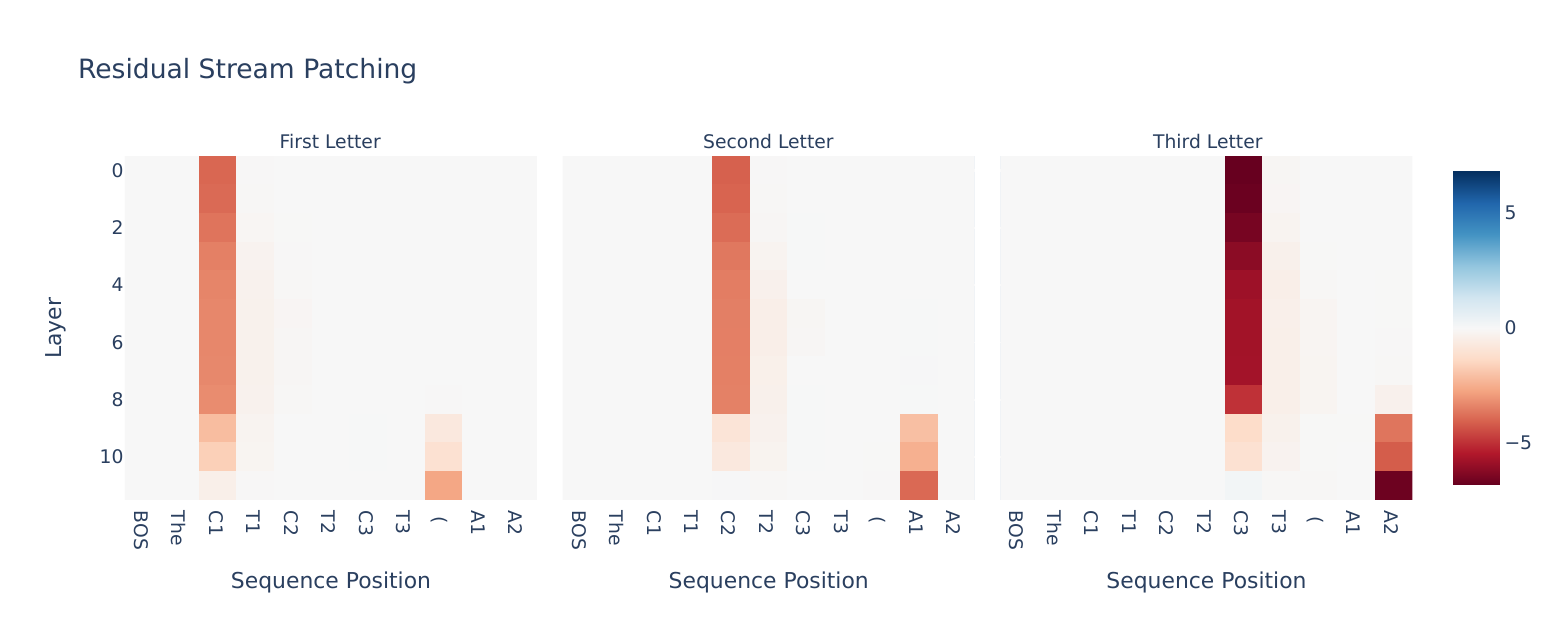}
    \caption{Patching the residual stream at every position and before every layer (corrupting the current word).}
    \label{fig:res_1}
\end{figure}

Fig. \ref{fig:res_1} shows the change in logit difference when patching the residual stream before each layer at every position, for each of the three letters of the acronym to predict. If patching the residual stream at a given position and layer considerably degrades the performance, then it implies that the activations stored at that specific step are important for the acronym prediction task. In this case, we can see that patching \texttt{Ci} at earlier layers does indeed drastically degrade the performance, with the logit difference dropping up to -5 units. We can also notice a shift from \texttt{Ci} to \texttt{A(i-1)} beginning at layer 8. This implies that there are some components that read information about \texttt{Ci}, move it to \texttt{A(i-1)} and then use it to perform the prediction of the next letter \texttt{Ai}.

\begin{figure}[htbp]
    \centering
    \includegraphics[width=\linewidth]{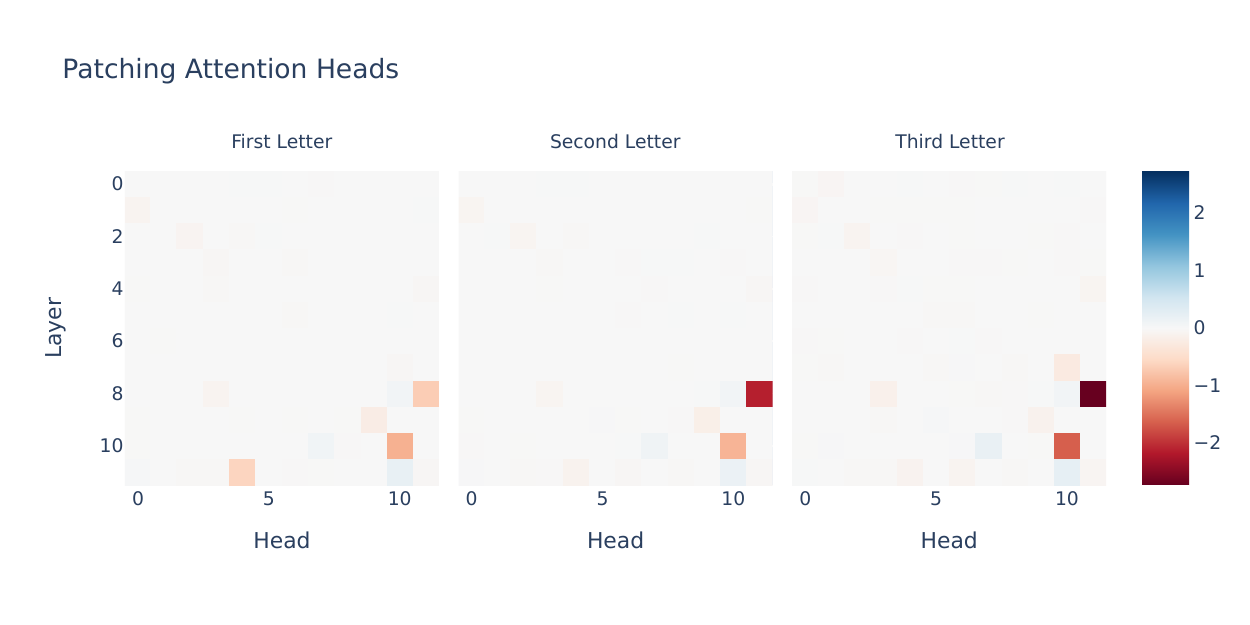}
    \caption{Patching the output of attention heads for every iteration (corrupting the current word).}
    \label{fig:attn_1}
\end{figure}

Once that we have tracked the flow of information, we can have a more fine-grained view by patching at the level of individual components, i.e. attention heads or MLPs. We performed activation patching experiments on MLPs and found that they were not relevant for acronym prediction. This was expected, as this task mostly requires moving information between token positions, which can only be performed by attention heads. Fig. \ref{fig:attn_1} shows the result of patching the output of attention heads with the activations obtained by corrupting the current word. We are able to localize four heads that are relevant across the three predictions: \texttt{8.11}, \texttt{10.10}, \texttt{9.9} and \texttt{11.4}. It is also interesting to notice that the drop on performance is larger on the last letter than on the first. This is due to the fact that the model has more context (i.e. the two previous letters of the acronym) when predicting the last letter, which translates to the model being more confident on its prediction, which corresponds to a larger drop when patching.

\begin{figure}[htbp]
    \centering
    \includegraphics[width=\linewidth]{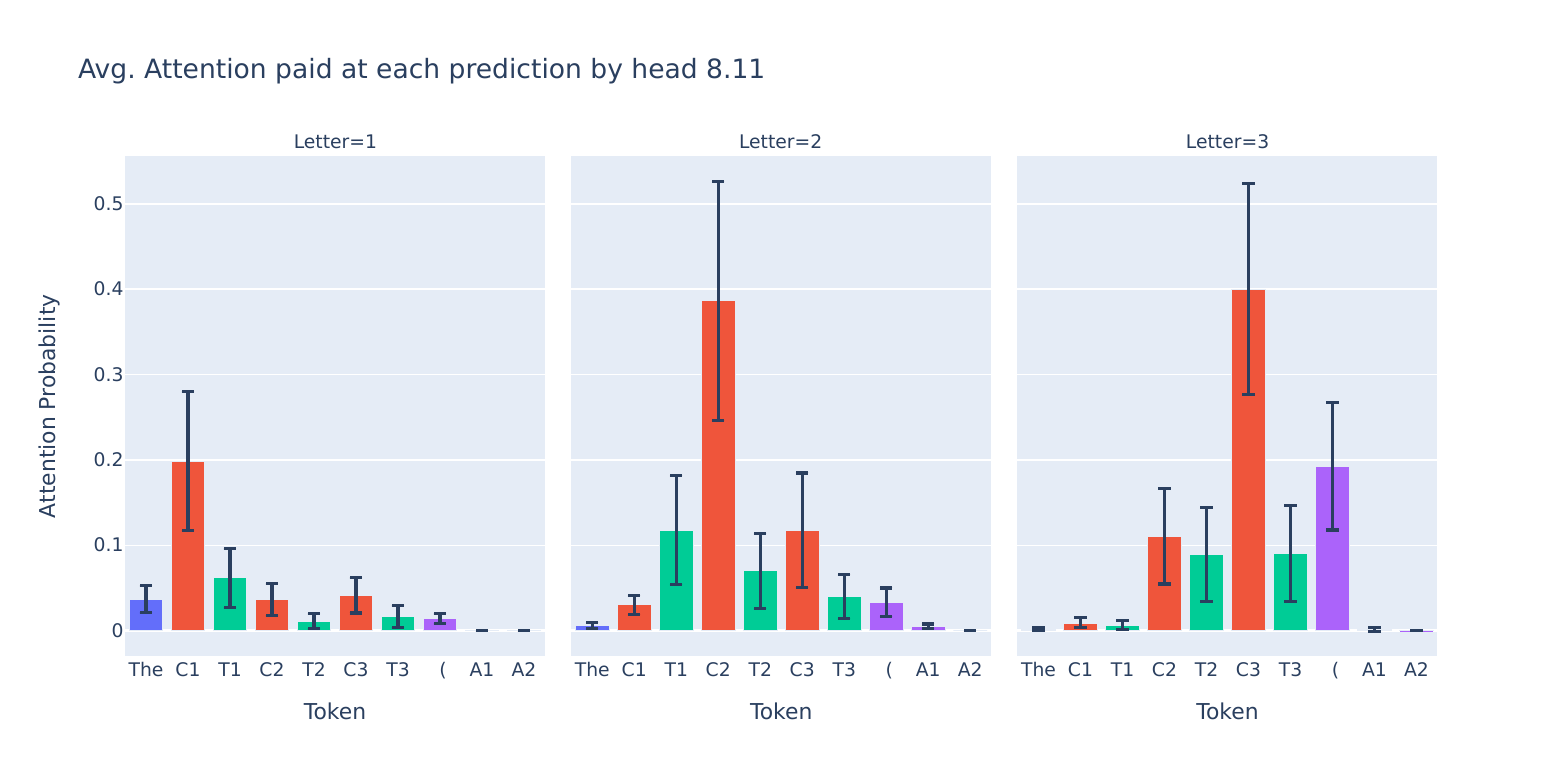}
    \caption{Average probability paid from \texttt{A(i-1)} to the previous token positions for head \texttt{8.11}.}
    \label{fig:baseline_histogram}
\end{figure}

Fig. \ref{fig:baseline_histogram} shows the distribution of the average attention paid from \texttt{A(i-1)} to the previous token positions for head \texttt{8.11}. It can clearly be seen that it mostly attends from \texttt{A(i-1)} to \texttt{Ci}, strongly suggesting that these heads copy the information of the corresponding letter and use it to perform the prediction of the next letter of the acronym, so we term this heads as \emph{letter mover heads}. The behavior of these heads will be extensively discussed on Section \ref{sec:LMH}, after performing the remaining activation patching experiments. The attention patterns for the rest of letter mover heads can be seen in the Supplementary Materials.

\subsubsection{Corrupting the Previous Words}

\begin{figure}[htbp]
    \centering
    \includegraphics[width=\linewidth]{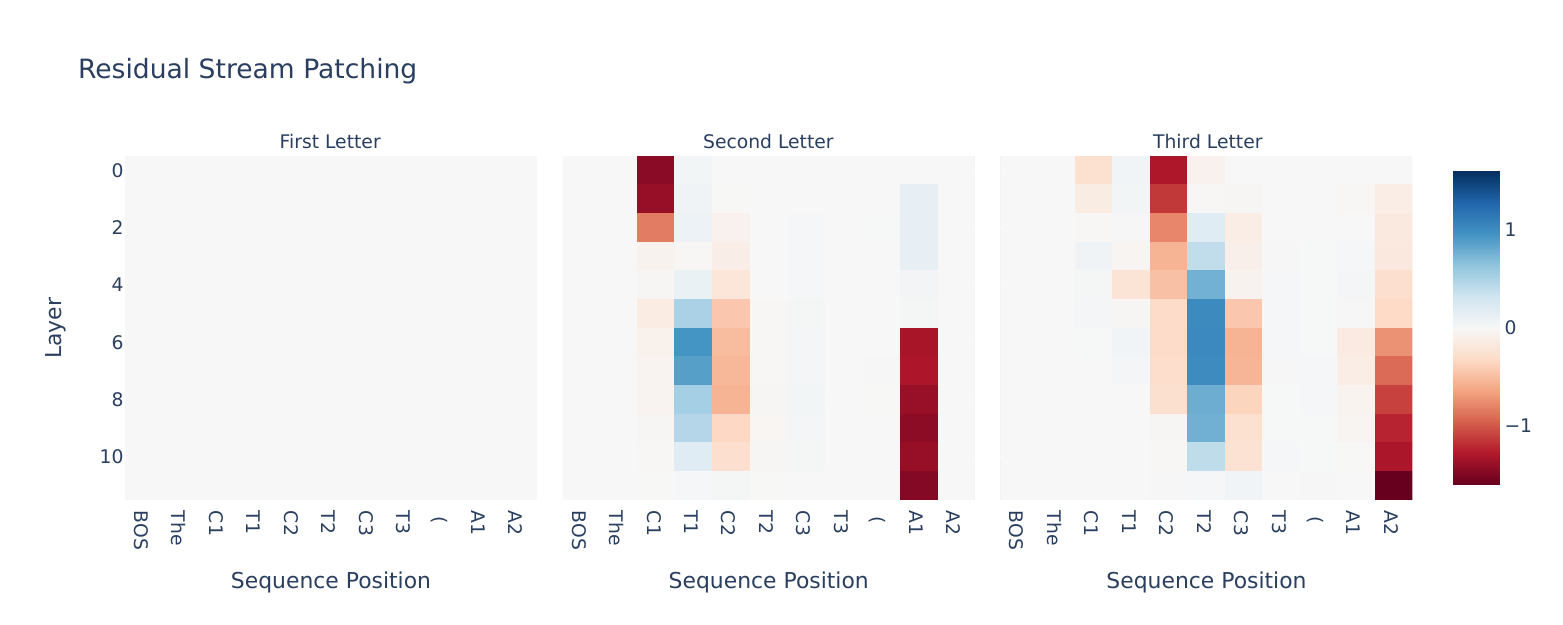}
    \caption{Patching the residual stream (corrupting previous words).}
    \label{fig:res_2}
\end{figure}

Fig. \ref{fig:res_2} shows the results of performing activation patching on the residual stream by corrupting the previous words. As expected, there is no effect on the prediction of the first letter because there are no previous words to corrupt. However, on the remaining letters, patching \texttt{C(i-1)} at earlier layers has a significant (although quite smaller than the previous) effect on predicting \texttt{Ai}, indicating that the circuit uses information about the previous words to perform the task. Concretely, it seems that the information is moved from \texttt{C(i-1)} to \texttt{Ci} on layers 1-2 and from there to \texttt{A(i-1)} at layer 5 and below. Interestingly, patching \texttt{Ti} around layers 5-11 slightly improves the performance of the circuit. We hypothesize that patching only this position may cause the model to become less confident on previous letters, essentially increasing the logit difference. As it is not the focus of the paper and the effect is minimal, we will not delve deeper into this fact.

\begin{figure}[htbp]
    \centering
    \includegraphics[width=\linewidth]{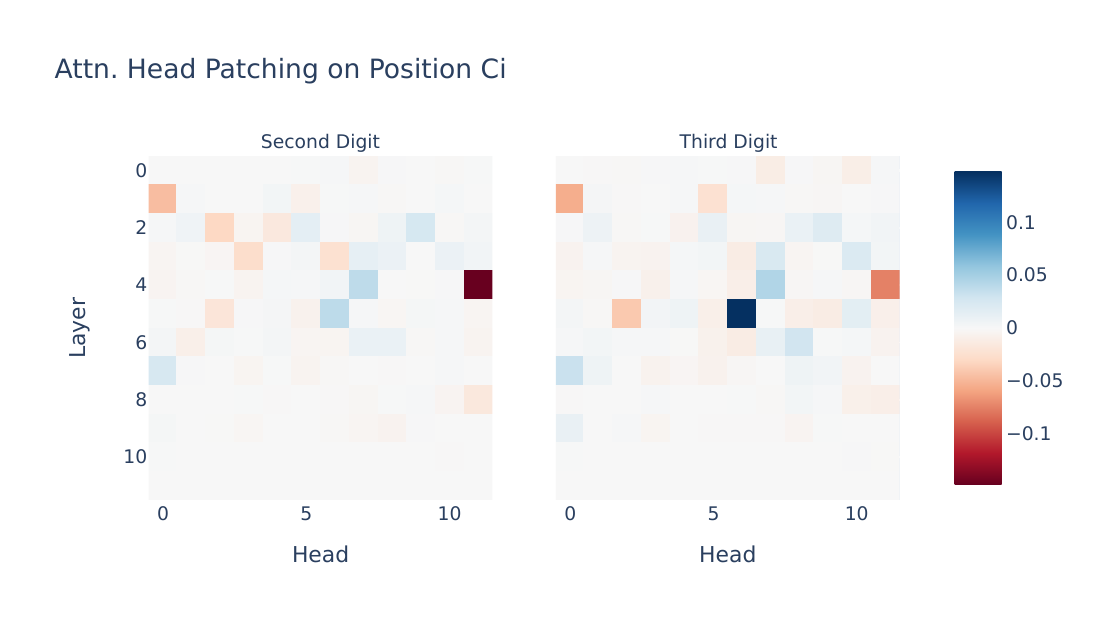}
    \caption{Patching the output of attention heads for every iteration at position \texttt{Ci} (corrupting the previous words).}
    \label{fig:attn_2_Ci}
\end{figure}

In order to check which attention heads were responsible for this movement of information, we patched the output of attention heads at positions \texttt{Ci} and \texttt{A(i-1)}. Fig. \ref{fig:attn_2_Ci} shows that there are a diffuse set of attention heads responsible for moving information from \texttt{C(i-1)} to \texttt{Ci}, such as \texttt{4.11}, \texttt{1.0} and \texttt{2.2}. Further inspection of the attention patterns show that they are \emph{previous token heads} (or fuzzy versions of it), i.e. heads that attend to the previous token position w.r.t. the current token and move information. Visualizations of the attention patterns of these heads can be found on the Supplementary Materials.

\begin{figure}[htbp]
    \centering
    \includegraphics[width=\linewidth]{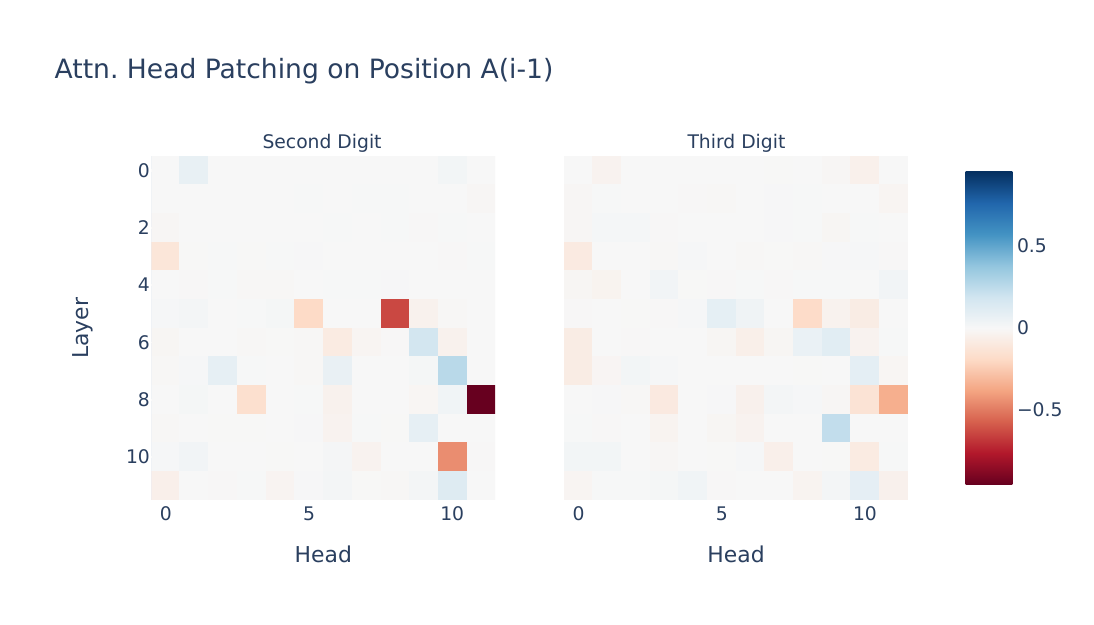}
    \caption{Patching the output of attention heads for every iteration at position \texttt{A(i-1)} (corrupting the previous words).}
    \label{fig:attn_2_Ai_1}
\end{figure}

On the other hand, Fig. \ref{fig:attn_2_Ai_1} shows that heads \texttt{5.8}, \texttt{8.11} and \texttt{10.10} are the most relevant in this patching experiment. Further inspection of the attention patterns reveals that they mostly attend to the \texttt{T(i-1)} and \texttt{Ci} tokens and move the information that was propagated to these positions via the previous token heads. 

There are a few important aspects to remark from this patching experiment. The first is that the performance drop when patching individual components is significantly lower than on the previous experiment. Secondly, the computation is more diffuse, i.e. it is distributed across many components, specially when looking at the \texttt{Ci} position. As we will see in the next section, this is due to the fact that the model is able to obtain the exact same information (i.e. the capital letter of the previous word) by just attending to the previous predicted letter, which is considerably easier. Another interesting fact is that some letter mover heads are also present in this part of the computation, i.e. some heads have multiple roles or behaviors, which is a motif that has also been discovered on other works \cite{docstring}.

\subsubsection{Corrupting Previous Predicted Letters}

\begin{figure}[htbp]
    \centering
    \includegraphics[width=\linewidth]{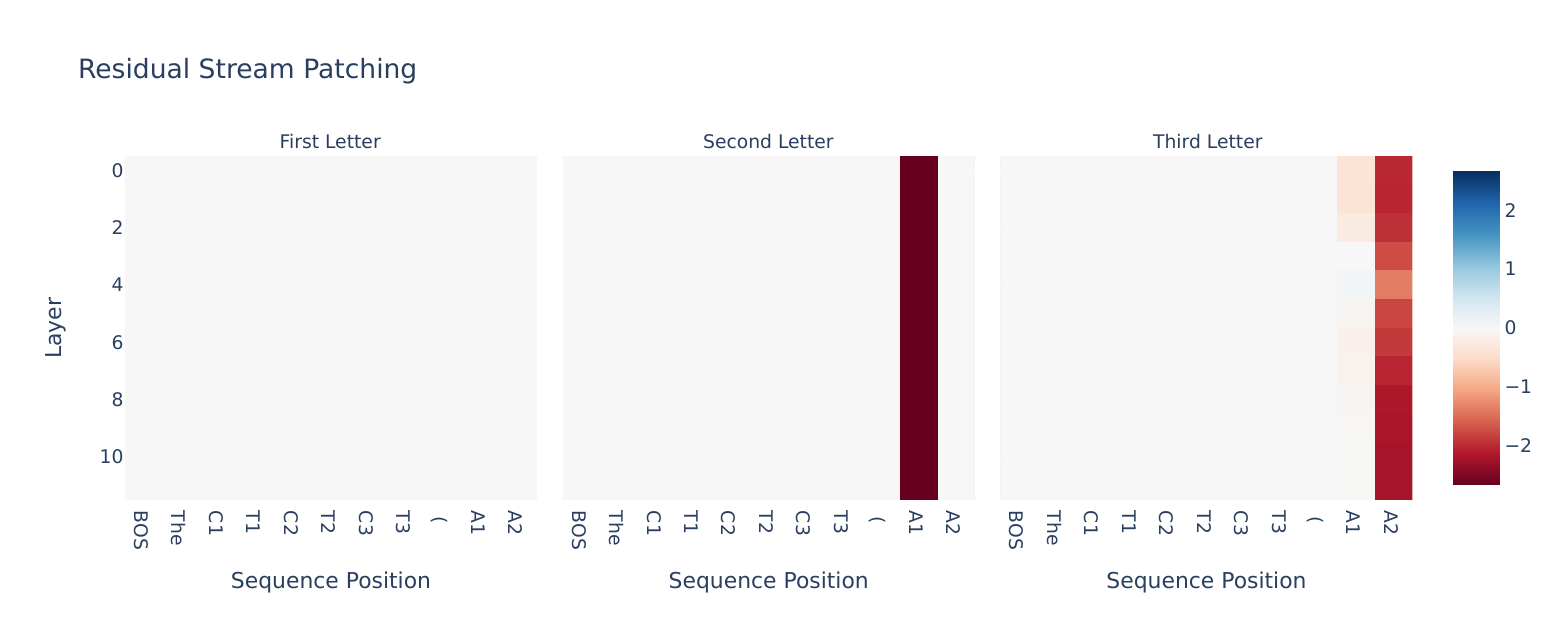}
    \caption{Patching the residual stream at every position and before every layer (corrupting the previous predicted letters).}
    \label{fig:res_3}
\end{figure}

The results presented in Fig. \ref{fig:res_3} clearly show that the model uses information about the previous predicted letter to predict the next one, as patching the \texttt{A(i-1)} position causes a considerable performance drop (larger than the previous corruption method) across every layer. This provides even more evidence in favor of the previously presented hypothesis that letter mover heads obtain the same information via two paths: from \texttt{C(i-1)} via the combination of previous token heads and heads that move information to \texttt{Ci} and then to \texttt{A(i-1)}, and directly from \texttt{A(i-1)}. 

To summarize, we have been able to discover the following circuit via a series of activation patching experiments:

\begin{itemize}
    \item Heads \texttt{8.11}, \texttt{10.10}, \texttt{9.9} and \texttt{11.4}, termed Letter Mover Heads, attend mostly to the \texttt{Ci} token position from the \texttt{A(i-1)} token position and are the main responsible for acronym prediction on GPT-2.
    \item Letter Mover Heads use the previous predicted letter to attend to the correct token position and predict the next letter of the acronym.
    \item This information, although more faintly, is also propagated from \texttt{C(i-1)} to \texttt{Ci} via a set of fuzzy previous heads such as \texttt{4.11}, \texttt{1.0} and \texttt{2.2}, which is then moved from \texttt{Ci} to \texttt{A(i-1)} via heads \texttt{5.8}, \texttt{8.11} and \texttt{10.10}.
\end{itemize}

\subsection{Circuit Evaluation}

\begin{figure}[htbp]
    \centering
    \includegraphics[width=\linewidth]{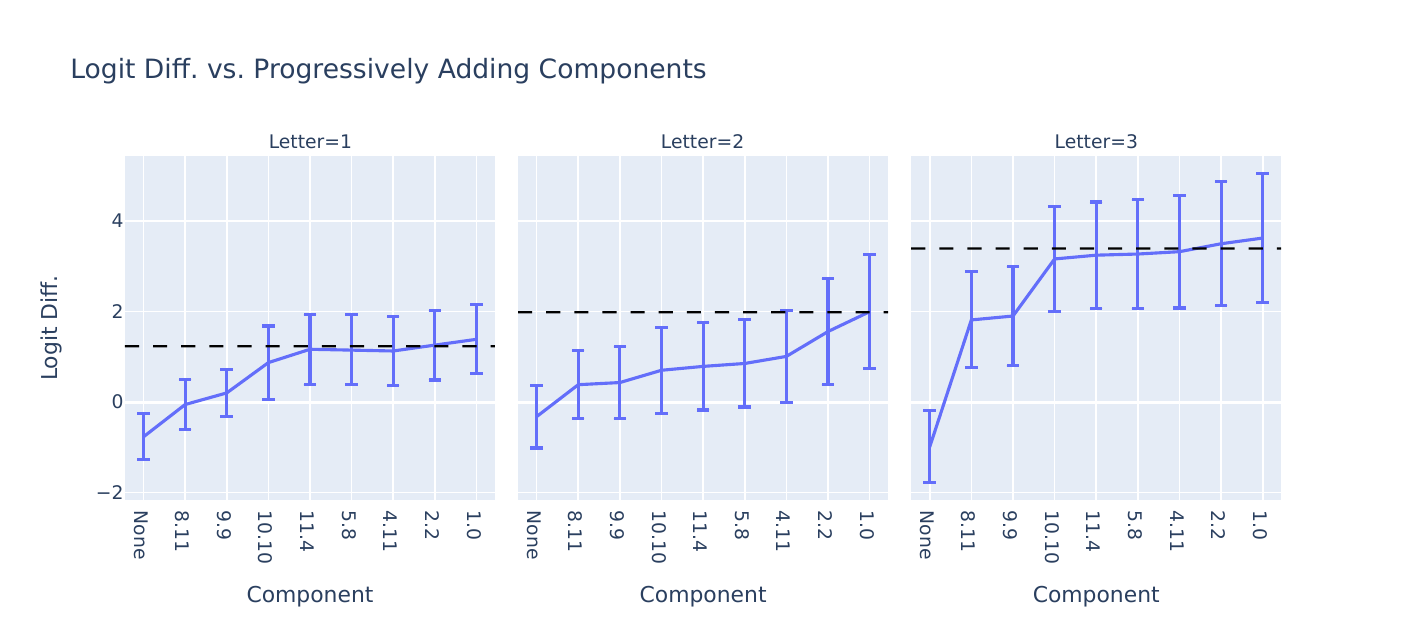}
    \caption{Logit Difference obtained by ablating everything and progressively adding components to the circuit. The dashed horizontal line represents the logit difference obtained with the complete model.}
    \label{fig:evaluation}
\end{figure}

Now that we have defined a circuit, it is necessary to evaluate whether it is sufficient to effectively perform acronym prediction. In order to evaluate it, we will ablate every other component that is not part of the circuit. Specifically, we will perform mean-ablation, which consists on replacing the activation of a component with the mean activation obtained across all samples of the dataset. In this way, we only discard the information related to the task of study while keeping the rest. Fig. \ref{fig:evaluation} shows the logit difference obtained when progressively adding heads to the circuit. We start with an empty circuit (i.e. ablating every head) to check that the model is unable to perform the task, obtaining negative values of the logit difference, as expected. Then, progressively adding Letter Mover Heads greatly improves performance, the most significant increase being on the third letter prediction, where adding just head \texttt{8.11} increases the average logit difference from -1 to 2 approximately. The logit difference keeps increasing by progressively adding the rest of components until we reach the baseline performance with just the 8 discovered heads. 
% In fact, it can be seen that the performance is even slightly higher than baseline, specially on the third letter prediction. 

\section{UNDERSTANDING LETTER MOVER HEADS}\label{sec:LMH}

Now that we have discovered and evaluated the main circuit responsible for the task of three-letter acronym prediction on GPT-2, we will provide further evidence on how Letter Mover Heads work, which are the main components of the circuit.

\subsection{What do Letter Mover Heads Copy?}

We discovered that Letter Mover Heads mostly attend from \texttt{A(i-1)} to \texttt{Ci} and were the main responsible for the acronym prediction task. Because of this, we hypothesize that these heads directly increase the logits of the correct letter to predict. In order to give evidence about this, we will take a look at the weights of Letter Mover Heads and try to reverse-engineer their behavior.

\begin{figure}[htbp]
    \centering
    \includegraphics[width=\linewidth]{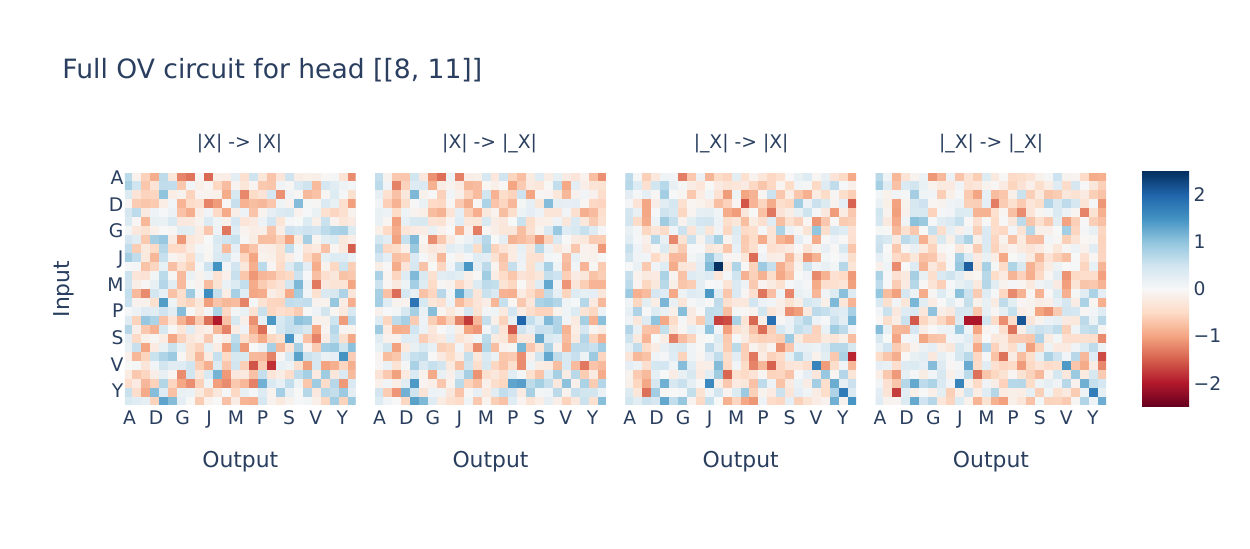}
    \caption{Full OV circuit for head \texttt{8.11}, for all capital letter tokens with/without a preceding space.}
    \label{fig:OV_1}
\end{figure}

Specifically, we will inspect the full OV circuit, obtained by retrieving the embeddings corresponding to the capital letter tokens and the capital letter tokens preceded by a space, passing them through the OV circuit of a Letter Mover Head and unembedding the resulting vector. This essentially tells us what would the head write into the residual stream if it attended perfectly to that token. Fig. \ref{fig:OV_1} shows the full OV circuit for Letter Mover Head \texttt{8.11}, rearranging it in four different ways to check what it writes when fully attending to capital letters, with or without a preceding space. At first sight, one cannot draw any conclusion except that there is a slight diagonal when attending to capital letters preceded with a space (two rightmost plots.

\begin{figure}[htbp]
    \centering
    \includegraphics[width=\linewidth]{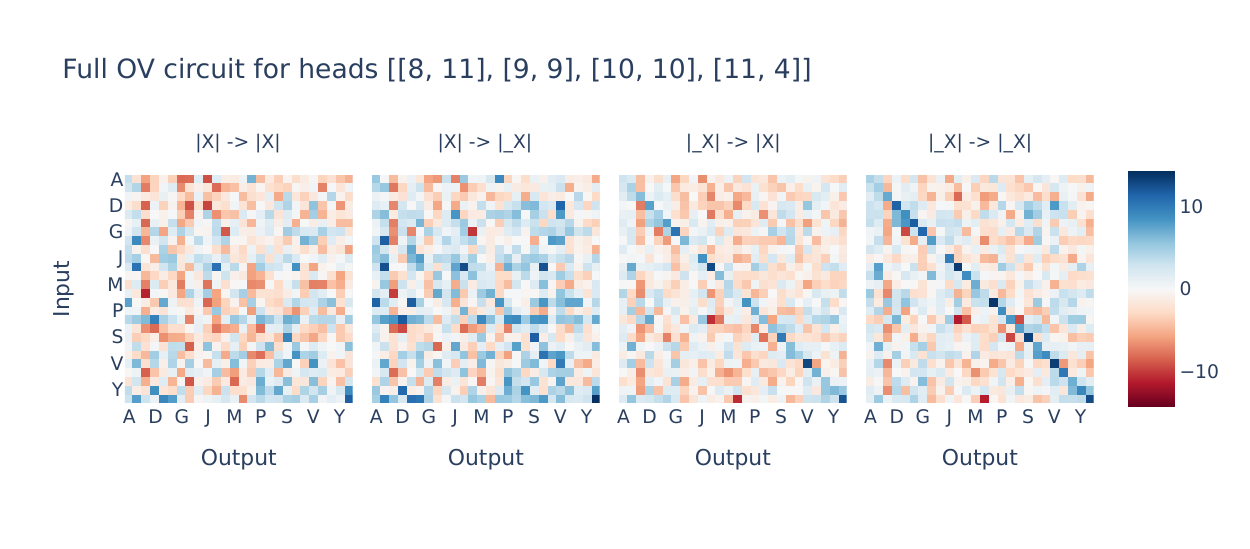}
    \caption{Sum of every Letter Mover Head OV circuit, for all capital letter tokens with/without a preceding space.}
    \label{fig:OV_4}
\end{figure}

However, the pattern becomes much more clear when we plot the full OV circuit taking into account all Letter Mover Heads. As it can be seen in Fig. \ref{fig:OV_4}, there is now a clear diagonal pattern on the two rightmost plots. In other words, this implies that when Letter Mover Heads attend to a capital letter preceded with a space (which is exactly what \texttt{Ci} are), they translate it to the token corresponding to the same capital letter without a space (i.e. \texttt{Ai}) and write it into the residual stream. It is important to remark that during analysis we did not use any information from the dataset, i.e. it was purely performed by inspecting the weights.

\begin{figure}[htbp]
    \centering
    \includegraphics[width=\linewidth]{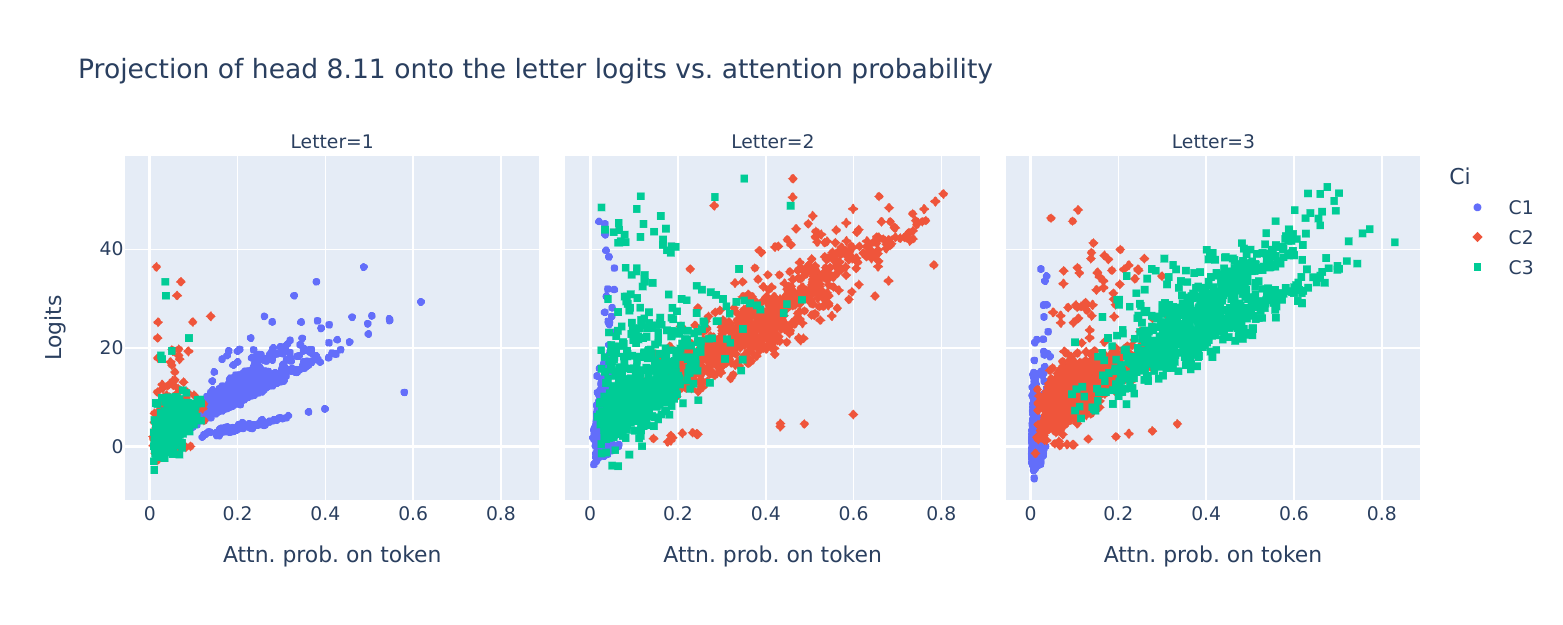}
    \caption{Projection of the output of head 8.11 along the logits correct letter vs. the attention probability paid to \texttt{Ci}.}
    \label{fig:scatter_8_11}
\end{figure}

We also studied the copying behavior by analyzing the relationship between the attention paid to \texttt{Ci} and the increase of the logits of \texttt{Ai} for different heads. Specifically, it can be seen on Fig. \ref{fig:scatter_8_11} that it pays more attention to \texttt{Ci} when predicting the $i$th digit, and that the value written along the logits of \texttt{Ai} increases with such attention. 

\subsection{Positional Information Experiments}

\begin{figure}[htbp]
    \centering
    \includegraphics[width=\linewidth]{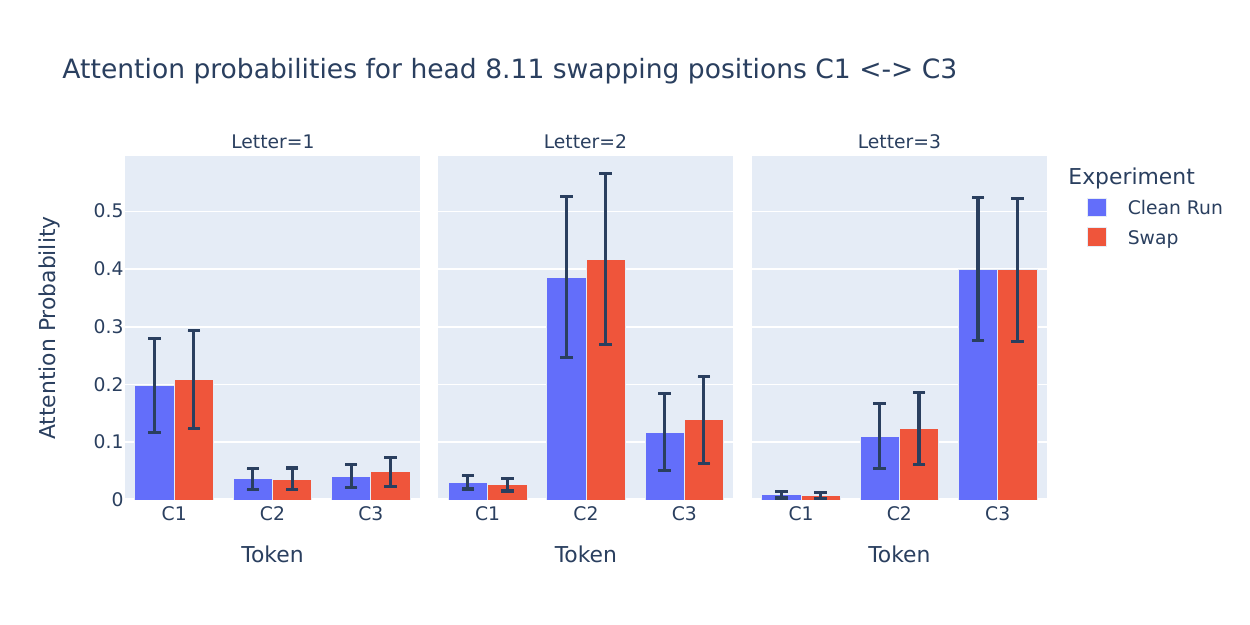}
    \caption{Attention probabilities on head \texttt{8.11} when swapping the positional embeddings of \texttt{C1} and \texttt{C3}.}
    \label{fig:swap_pos_C1_C3_8_11.pdf}
\end{figure}

We also hypothesized that Letter Mover Heads should use positional information to perform the final prediction (i.e. to attend to the first token of the \emph{first/second/third} word), specially when predicting the first letter of the acronym, as there is no available information regarding the previously predicted letters of the acronym. In order to test this hypothesis, we first study the positional embeddings, as it is the most evident source of positional information of the model. Specifically, we swapped the positional embeddings of different pairs of \texttt{Ci} and checked if it had an effect on the attention pattern of Letter Mover Heads. Ideally, if a head relied in positional embeddings, swapping them should force them to attend a different letter. Fig. \ref{fig:swap_pos_C1_C3_8_11.pdf} shows the effect of swapping the positional embeddings of \texttt{C1} and \texttt{C3} on the attention probabilities of head \texttt{8.11}. 

As it can be seen, the change in attention probabilities is negligible. We performed all the possible swapping combinations for every letter mover head and found similar results, implying that positional embeddings are not the main source of positional information for Letter Mover Heads. However, the model has to use some source of positional information to predict the first letter, so we looked for another possible source. It has been recently hypothesized \citep{docstring} that models are able to derive positional information from attention probabilities. Specially, due to causal masking, the attention pattern paid to the Beggining of Sequence (\texttt{BOS}) token position generally decreases with the destination token position. Therefore, the model could infer the position of a certain token \texttt{Ci} by looking at the attention paid to the \texttt{BOS} token: a lower attention paid to this token position implies that the destination token is further from the start of the sentence, and vice versa.

\begin{figure}[htbp]
    \centering
    \includegraphics[width=\linewidth]{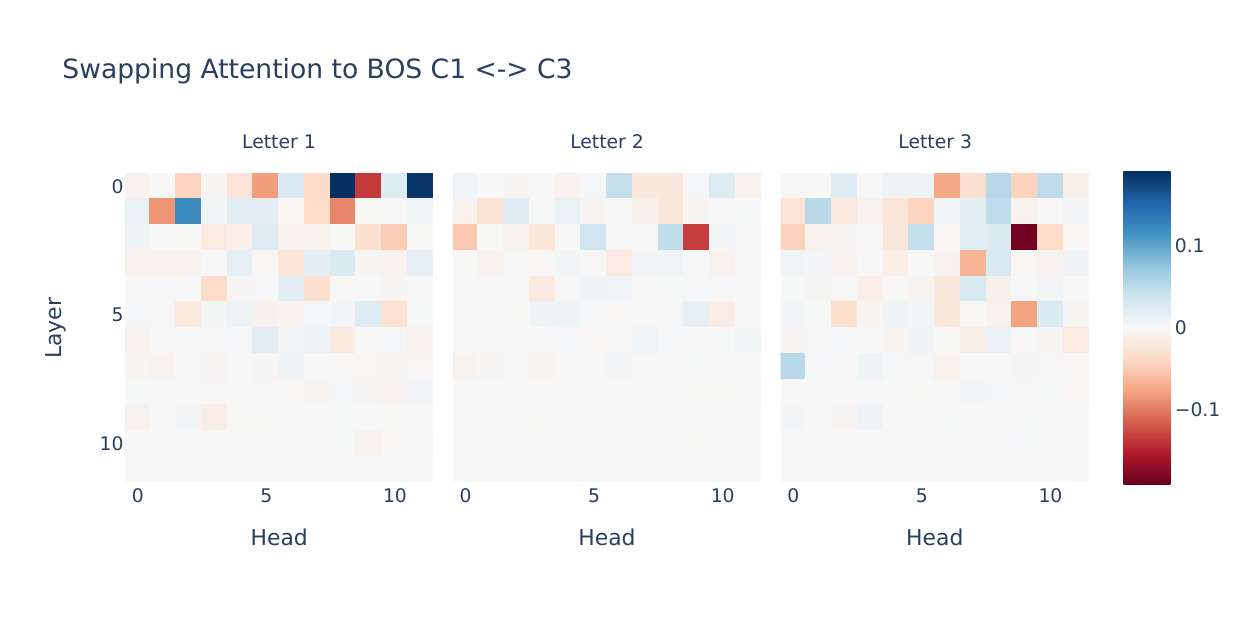}
    \caption{Change in logit difference obtained by swapping the attention paid to \texttt{BOS} from the \texttt{C1} and \texttt{C3} for every head in the model.}
    \label{fig:swap_BOS_C1_C3_attn_patching}
\end{figure}

Therefore, we patched the activations of each head by swapping their attention paid to the \texttt{BOS} token from tokens \texttt{Ci} and \texttt{Cj} for all possible combinations and measured the change in logit difference. Fig. \ref{fig:swap_BOS_C1_C3_attn_patching} shows the results for $i=1$, $j=3$. Indeed, swapping the attention values does have an impact on the performance, meaning that letter mover heads are likely to use positional information derived from this mechanism, specially when predicting the first letter. There are also other heads that contribute positively. We hypothesize that these heads are writing on the opposite direction to avoid the model becoming overconfident (similar to negative name mover heads on \cite{wang2022interpretability}). However, we leave this aspect as part of a future study, as this requires an extensive analysis.

\begin{figure}[htbp]
    \centering
    \includegraphics[width=\linewidth]{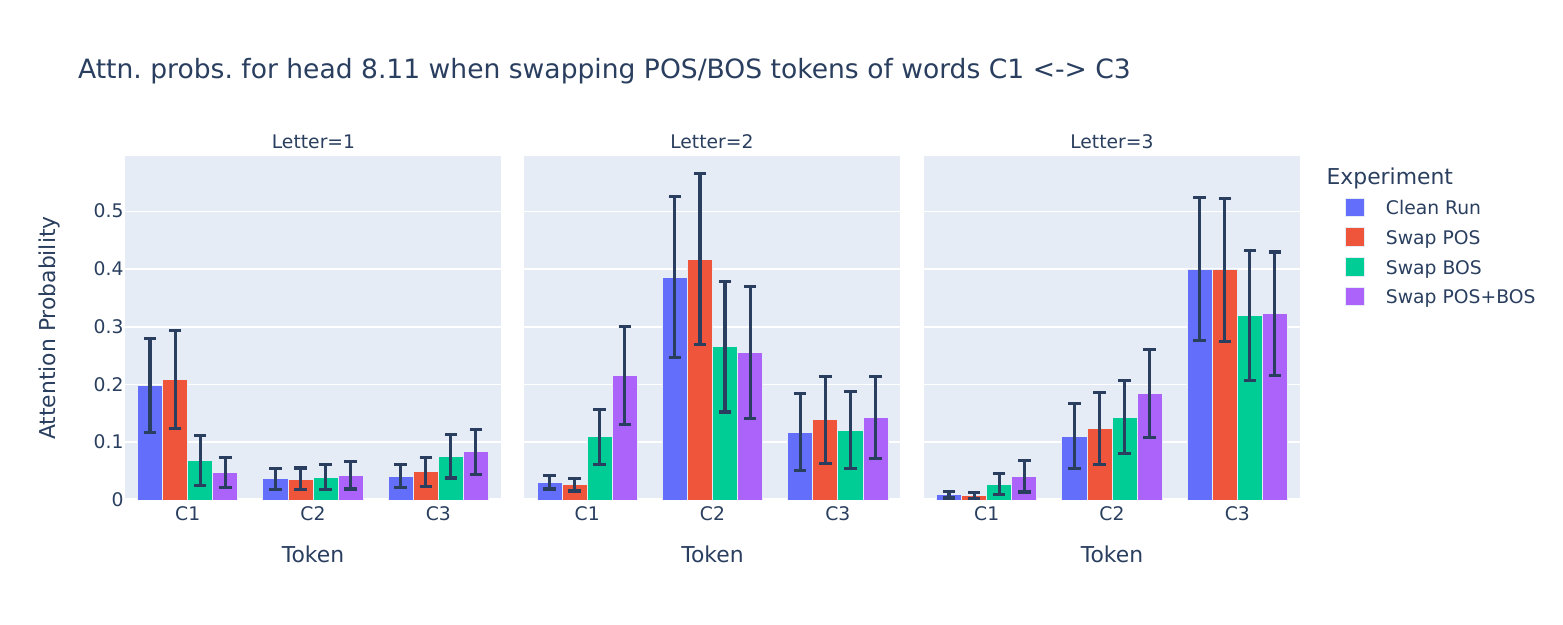}
    \caption{Change in attention when performing the \texttt{BOS} attention swapping experiment.}
    \label{fig:swap_BOS_C1_C3_8_11}
\end{figure}

In order to provide further evidence, we swapped the \texttt{BOS} attentions for those heads that had a negative impact of at least $1\%$ in the previous experiment and visualized the change of attention pattern on letter mover heads. Specifically, Fig. \ref{fig:swap_BOS_C1_C3_8_11} shows the attention probabilities paid to the \texttt{Ci} tokens on head \texttt{8.11} on the clean run, swapping the positional embeddings, swapping the \texttt{BOS} tokens and applying both swapping techniques. In general, swapping the \texttt{BOS} tokens has the most impact across all predictions, meaning that head \texttt{8.11} does indeed use positional information. As expected, the greatest difference can be found on predicting the first letter: swapping the positional embeddings and the \texttt{BOS} tokens of \texttt{C1} and \texttt{C3} changes the average prediction from \texttt{A1} to \texttt{A3}. It is also important to remark that, most of the change on the attention pattern, is caused by simply swapping two scalars on each of the patched heads, i.e. a slight change in the attention pattern of the patched heads causes a large impact on the attention probabilities of head \texttt{8.11}. 

We also performed this experiment with every possible swapping combination and found two important aspects. First, the attention probabilities are mainly affected on the first letter prediction, aligned with our hypothesis that the model relies mainly on positional information (i.e. it has to look for the \emph{first} capital letter). Comparatively, the second and third letter prediction rely on the previous predicted letter/s to gain some context. Second, we found that only swapping \texttt{C1} and \texttt{C3} had a considerable impact, probably due to the fact that the other two possible swaps are performed between tokens that are closer together, hence the degree of corruption is smaller. This phenomena also occured on the rest of letter mover heads. The experiment involving the rest of letter mover heads and swapping combinations are presented in the Supplementary Materials.

% As positional information is mainly used on the first letter prediction, it was expected that swapping \texttt{C2} and \texttt{C3}, whereas swapping \texttt{C1} and \texttt{C2} may not show a large impact due to the fact that these tokens are only 2 positions apart

\section{CONCLUSIONS}\label{sec:conclusions}

In this work, we identified the circuit responsible for the task of predicting three-letter acronyms on GPT-2 Small via a series of activation patching experiments. The discovered circuit was composed by 8 attention heads which we classified into three different groups according to their role. We showed that ablating every other head did preserve the performance, meaning that the task of acronym prediction does indeed rely on the discovered circuit. We also paid special attention to the most important heads of the circuit, which we termed \emph{letter mover heads}, whose role is to attend to the capital letter of the $i$th word and copy its content for the $i$th letter prediction. We provided evidence of this behavior by studying their attention patterns, OV matrices and output activations. We also show that these heads use positional information and that this information is received not only by the positional embeddings, but from the attention patterns. Our experiments show that the positional information is derived from the attention paid to the \texttt{BOS} token, in accordance to what it was discovered in simpler models \citep{docstring}.

In summary, this is the first work that tries to mechanistically interpret a task involving multiple consecutive token using MI, laying the foundation for understanding more complex behaviors. Moreover, we strongly believe that MI will enable us to understand larger models, increasing the safety and trustworthiness of AI systems.

% \todo{PONER UN PAR DE CITAS DEL AISTATS}
% \todo{FECHA VISITA POSTS}

% Despite our work being limited to the study of a single model and task, we found mechanisms that were also discovered on other tasks/models, suggesting that we can find \emph{motifs} that are universal across neural networks. Concretely, we found that the behavior of letter mover heads is also found on name mover heads in \cite{wang2022interpretability} or \cite{docstring}. Also, our positional information experiments showed that the circuit was able to read positional information from the attention patterns. We think that this is a rich opportunity to understand models, hence more refined methods are required to further track down the positional information flow. As previously-mentioned, the model that we studied on this work is orders of magnitude smaller than the ones used in production. Even though the study of smaller models is both necessary and useful, efforts should be put into developing methods to understand a wider range of larger models. One of the possibilities would be to automate the process of circuit discovery \cite{conmy2023towards}.

% \todo{REVISAR FECHAS BIBLIOGRAFIA (POST)}

% \todo{Mencionar generalizabilidad, estamos trabajando en entender multiples tokens (establece la base) y se podra aplicar a procesos mas complejos.}

\subsubsection*{Acknowledgements}
This work has been co-funded by the BALLADEER (PROMETEO/2021/088) project, a Big Data analytical platform for the diagnosis and treatment of Attention Deficit Hyperactivity Disorder (ADHD) featuring extended reality, funded by the \emph{Conselleria de Innovación, Universidades, Ciencia y Sociedad Digital (Generalitat Valenciana)} and the AETHER-UA project (PID2020-112540RB-C43), a smart data holistic approach for context-aware data analytics: smarter machine learning for business modelling and analytics, funded by the \emph{Spanish Ministry of Science and Innovation}. Jorge García-Carrasco holds a predoctoral contract (CIACIF/2021/454) granted by the \emph{Conselleria de Innovación, Universidades, Ciencia y Sociedad Digital (Generalitat Valenciana)}.

\bibliographystyle{plainnat}
\bibliography{mybibliography}

\section*{Checklist}

% %%% BEGIN INSTRUCTIONS %%%
% The checklist follows the references. For each question, choose your answer from the three possible options: Yes, No, Not Applicable.  You are encouraged to include a justification to your answer, either by referencing the appropriate section of your paper or providing a brief inline description (1-2 sentences). 
% Please do not modify the questions.  Note that the Checklist section does not count towards the page limit. Not including the checklist in the first submission won't result in desk rejection, although in such case we will ask you to upload it during the author response period and include it in camera ready (if accepted).

% \textbf{In your paper, please delete this instructions block and only keep the Checklist section heading above along with the questions/answers below.}
% %%% END INSTRUCTIONS %%%

 \begin{enumerate}

 \item For all models and algorithms presented, check if you include:
 \begin{enumerate}
   \item A clear description of the mathematical setting, assumptions, algorithm, and/or model. [Yes/No/Not Applicable] Yes, at Section \ref{sec:background}.
   \item An analysis of the properties and complexity (time, space, sample size) of any algorithm. [Yes/No/Not Applicable] Not applicable. The focus of this work is not on any algorithm but on discovering a circuit by using an already existing technique. However, we specify the sample size used on Section \ref{sec:circuit}.  
   \item (Optional) Anonymized source code, with specification of all dependencies, including external libraries. [Yes/No/Not Applicable] No, but it will be made public upon acceptance.
 \end{enumerate}

 \item For any theoretical claim, check if you include:
 \begin{enumerate}
   \item Statements of the full set of assumptions of all theoretical results. [Yes/No/Not Applicable] Not applicable. The work presented here is mainly empirical evidence.
   \item Complete proofs of all theoretical results. [Yes/No/Not Applicable] Not applicable.
   \item Clear explanations of any assumptions. [Yes/No/Not Applicable] Not applicable.
 \end{enumerate}

 \item For all figures and tables that present empirical results, check if you include:
 \begin{enumerate}
   \item The code, data, and instructions needed to reproduce the main experimental results (either in the supplemental material or as a URL). [Yes/No/Not Applicable] No, but it will be made publicly available upon acceptance.
   \item All the training details (e.g., data splits, hyperparameters, how they were chosen). [Yes/No/Not Applicable] Not applicable, as no training is performed. We study a pretrained model.
 \item A clear definition of the specific measure or statistics and error bars (e.g., with respect to the random seed after running experiments multiple times). [Yes/No/Not Applicable] Yes
 \item A description of the computing infrastructure used. (e.g., type of GPUs, internal cluster, or cloud provider). [Yes/No/Not Applicable] Yes, on the start of Section \ref{sec:circuit}.
 \end{enumerate}

 \item If you are using existing assets (e.g., code, data, models) or curating/releasing new assets, check if you include:
 \begin{enumerate}
   \item Citations of the creator If your work uses existing assets. [Yes/No/Not Applicable] Yes
   \item The license information of the assets, if applicable. [Yes/No/Not Applicable] Not applicable
   \item New assets either in the supplemental material or as a URL, if applicable. [Yes/No/Not Applicable] Yes
   \item Information about consent from data providers/curators. [Yes/No/Not Applicable] Not applicable
   \item Discussion of sensible content if applicable, e.g., personally identifiable information or offensive content. [Yes/No/Not Applicable] Not applicable
 \end{enumerate}

 \item If you used crowdsourcing or conducted research with human subjects, check if you include:
 \begin{enumerate}
   \item The full text of instructions given to participants and screenshots. [Yes/No/Not Applicable] Not applicable
   \item Descriptions of potential participant risks, with links to Institutional Review Board (IRB) approvals if applicable. [Yes/No/Not Applicable] Not applicable
   \item The estimated hourly wage paid to participants and the total amount spent on participant compensation. [Yes/No/Not Applicable] Not applicable
 \end{enumerate}

 \end{enumerate}

%%%%%%%%%%%%%%%%%%%%%%%%%%%%%%%%%%%%%%%%%%%%%%%%%%%%%%%%%%%%
\clearpage

%\includepdf[pages=-]{supplement.pdf}

\onecolumn
%\aistatstitle{Supplementary Material for: How does GPT-2 Predict Acronyms? Extracting and Understanding a Circuit via Mechanistic Interpretability}
\appendix

\section{ATTENTION PATTERNS}

\begin{figure}[htbp]
    \centering
    \includegraphics[width=\linewidth]{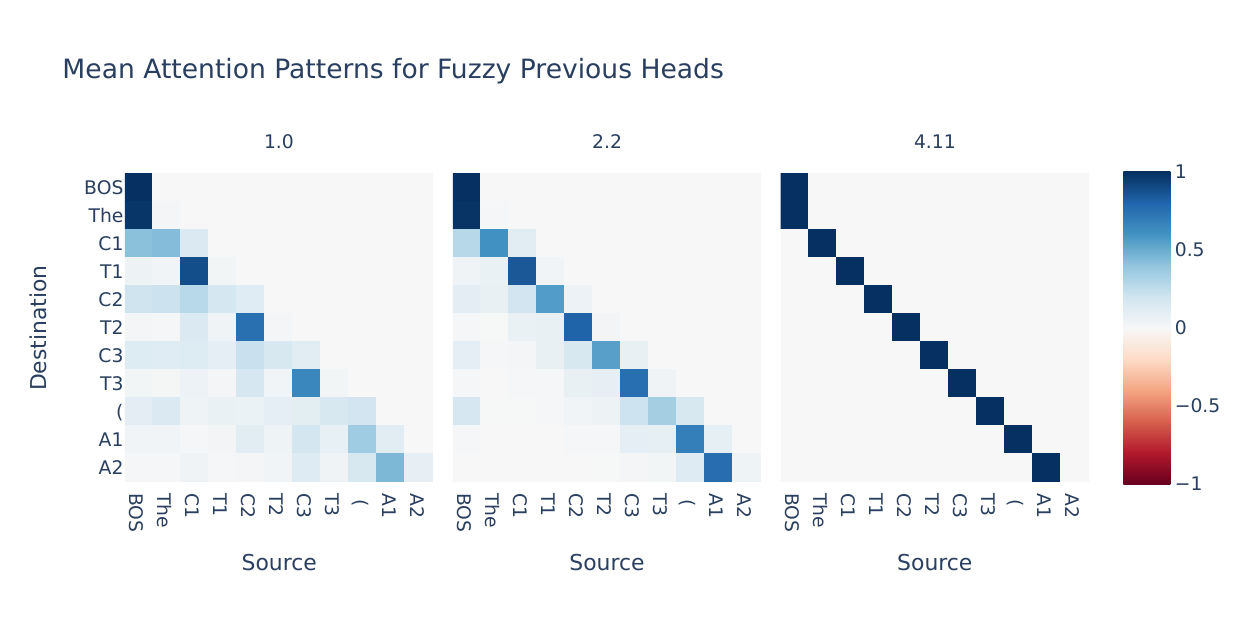}
    \caption{Mean attention patterns for the 3 heads on the circuit that move information from \texttt{C(i-1}) to \texttt{Ci}.}
    \label{fig:attn_patterns_fuzzy}
\end{figure}

\begin{figure}[htbp]
    \centering
    \includegraphics[width=\linewidth]{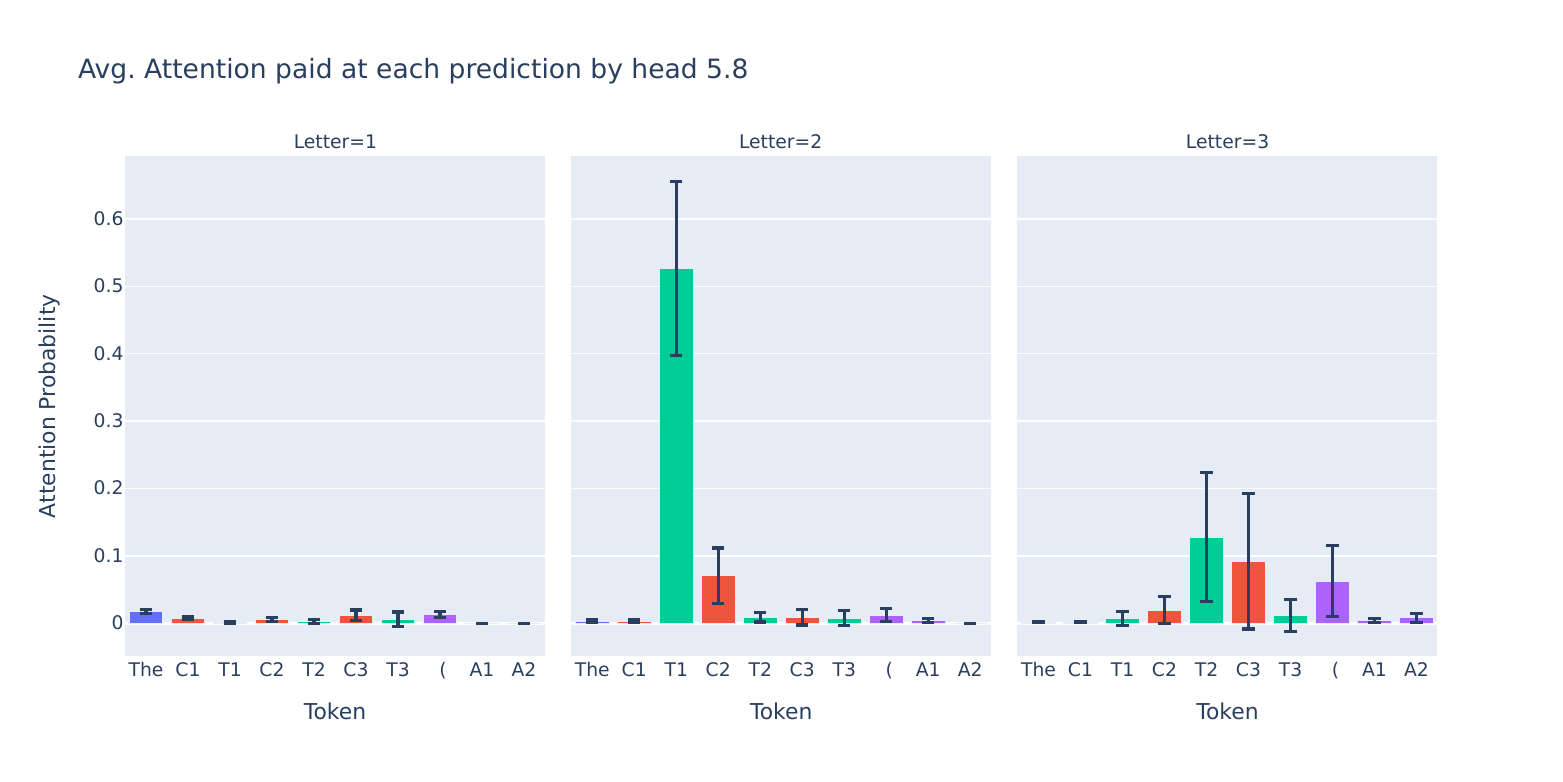}
    \caption{Average probability paid from \texttt{A(i-1)} to the
previous token positions for head \texttt{5.8}.}
    \label{fig:hist_5_8}
\end{figure}

\begin{figure}[htbp]
    \centering
    \includegraphics[width=\linewidth]{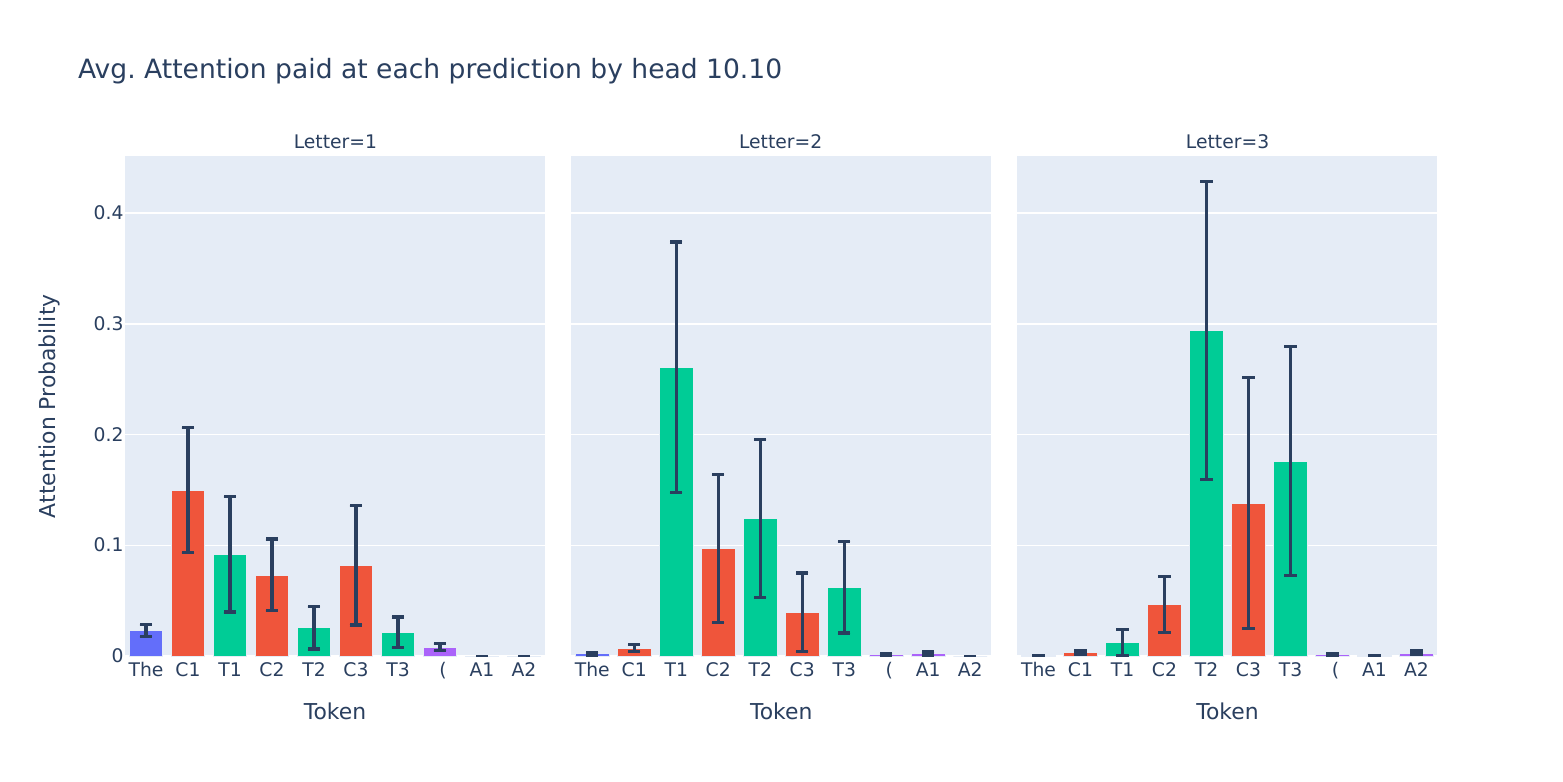}
    \caption{Average probability paid from \texttt{A(i-1)} to the
previous token positions for head \texttt{10.10}.}
    \label{fig:hist_10_10}
\end{figure}

This section contains additional attention pattern visualizations to support the findings described in Section 3. Fig. \ref{fig:attn_patterns_fuzzy} shows the mean attention patterns for heads \texttt{1.0}, \texttt{2.2} and \texttt{4.11}, which were the main responsible of moving information from previous words. As it can be seen, these heads have the characteristic offset diagonal pattern of previous token heads, meaning that these heads attend to the previous token w.r.t. the current one and copy their information. On the other hand, heads \texttt{5.8}, \texttt{8.11} and \texttt{10.10} move the previous information into \texttt{A(i-1)} by attending to the tokens \texttt{T(i-1)} and \texttt{Ci}, as it can be seen on Figs. \ref{fig:hist_5_8}-\ref{fig:hist_10_10}.

\begin{figure}[htbp]
    \centering
    \includegraphics[width=\linewidth]{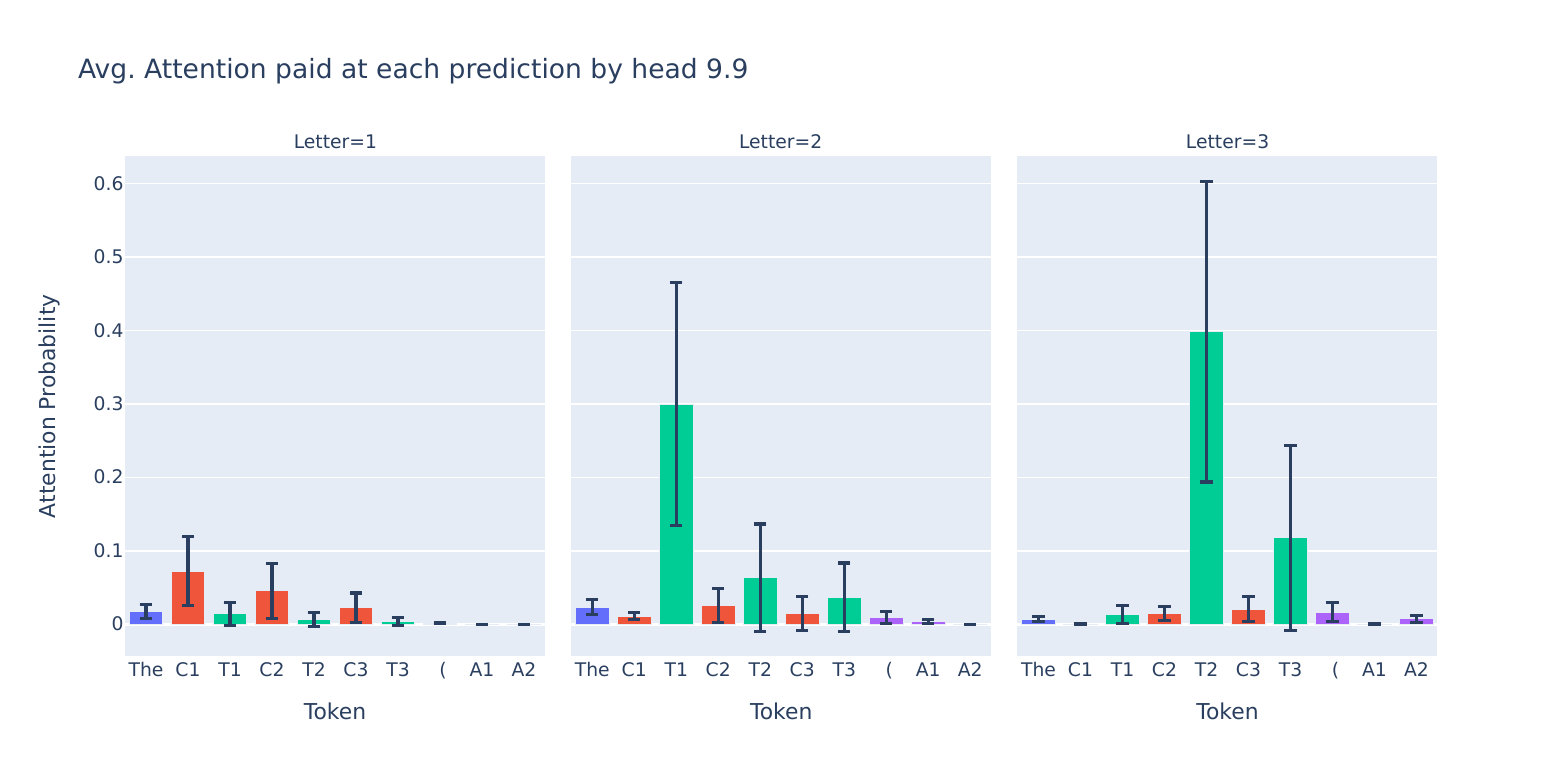}
    \caption{Average probability paid from \texttt{A(i-1)} to the
previous token positions for head \texttt{9.9}.}
    \label{fig:hist_9_9}
\end{figure}

\begin{figure}[htbp]
    \centering
    \includegraphics[width=\linewidth]{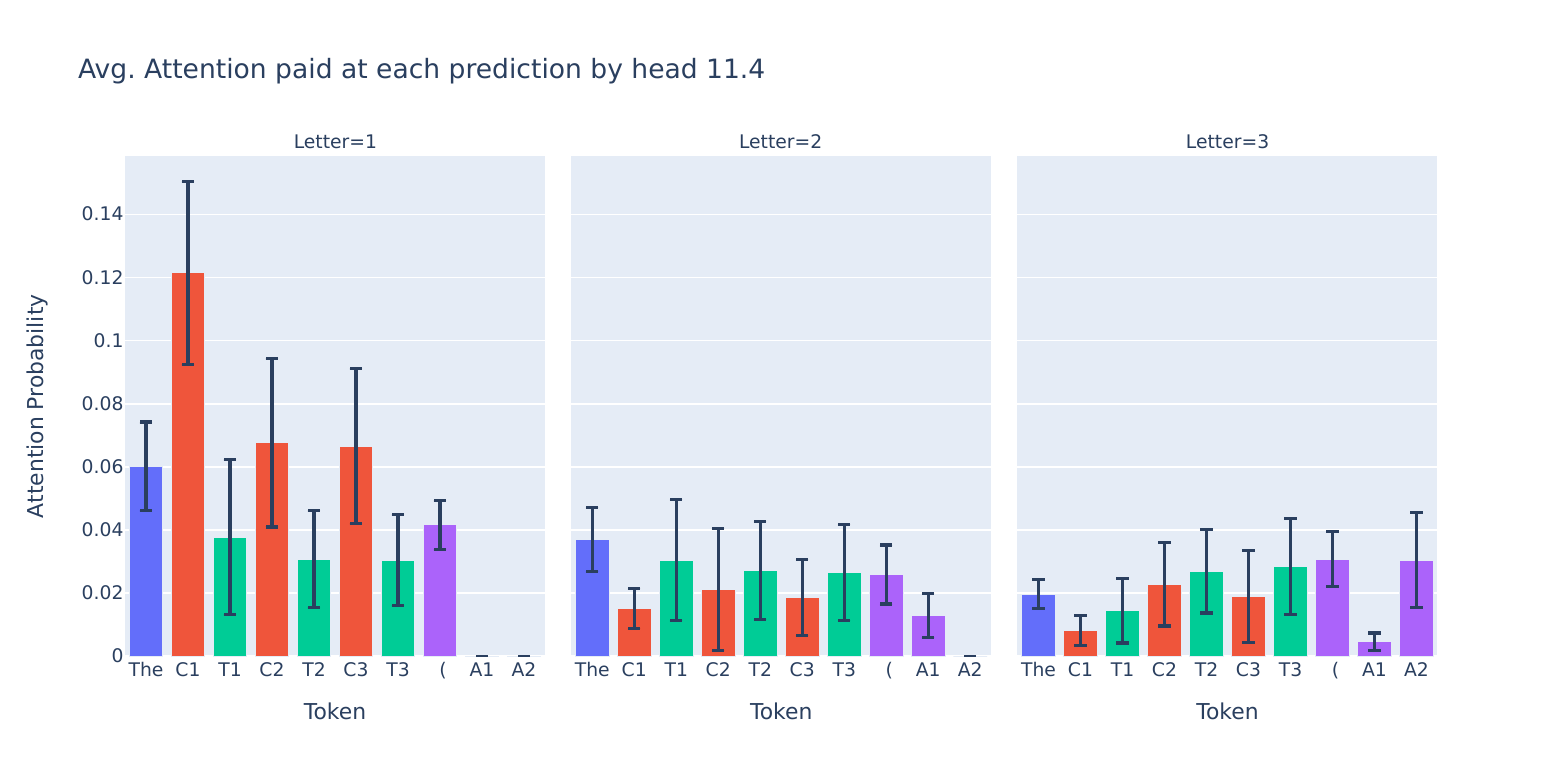}
    \caption{Average probability paid from \texttt{A(i-1)} to the
previous token positions for head \texttt{11.4}.}
    \label{fig:hist_11_4}
\end{figure}

The attention probability histograms of the rest of letter mover heads \texttt{10.10}, \texttt{9.9} and \texttt{11.4} are shown on Figs. \ref{fig:hist_10_10}-\ref{fig:hist_11_4}. Even though the histograms are noisier than the one associated to the main letter mover head \texttt{8.11}, it can clearly be seen that these heads generally pay more attention to the proper capital letter \texttt{Ci} compared with the other capital letters. We also see a high attention paid to \texttt{T(i-1)}, in particular on heads \texttt{9.9} and \texttt{10.10}. This is likely due to the fact that these heads (specially \texttt{10.10}) also perform the role of copying information about the previous capital letter, as previously-mentioned.

\vfill

\FloatBarrier

\section{POSITIONAL INFORMATION EXPERIMENTS}

\begin{figure}[htbp]
    \centering
    \includegraphics[width=\linewidth]{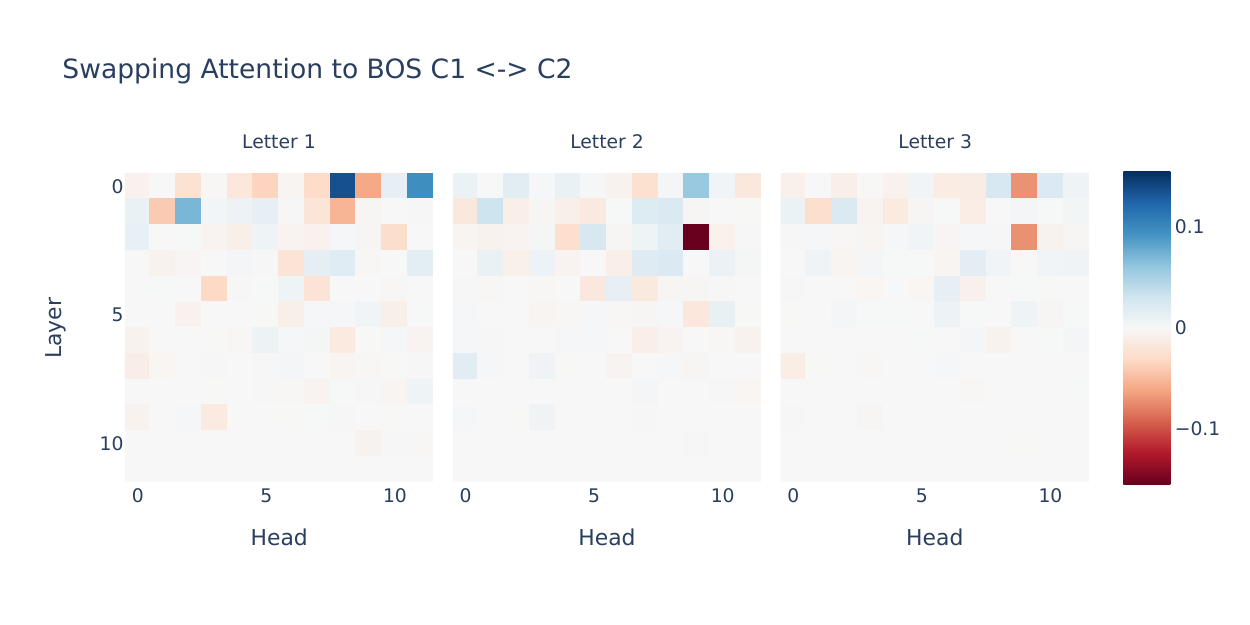}
    \caption{Change in logit difference obtained by swapping the attention paid to \texttt{BOS} from the \texttt{C1} and \texttt{C2} tokens for every head in the model.}
    \label{fig:swap_BOS_C1_C2_attn_patching}
\end{figure}

\begin{figure}[htbp]
    \centering
    \includegraphics[width=\linewidth]{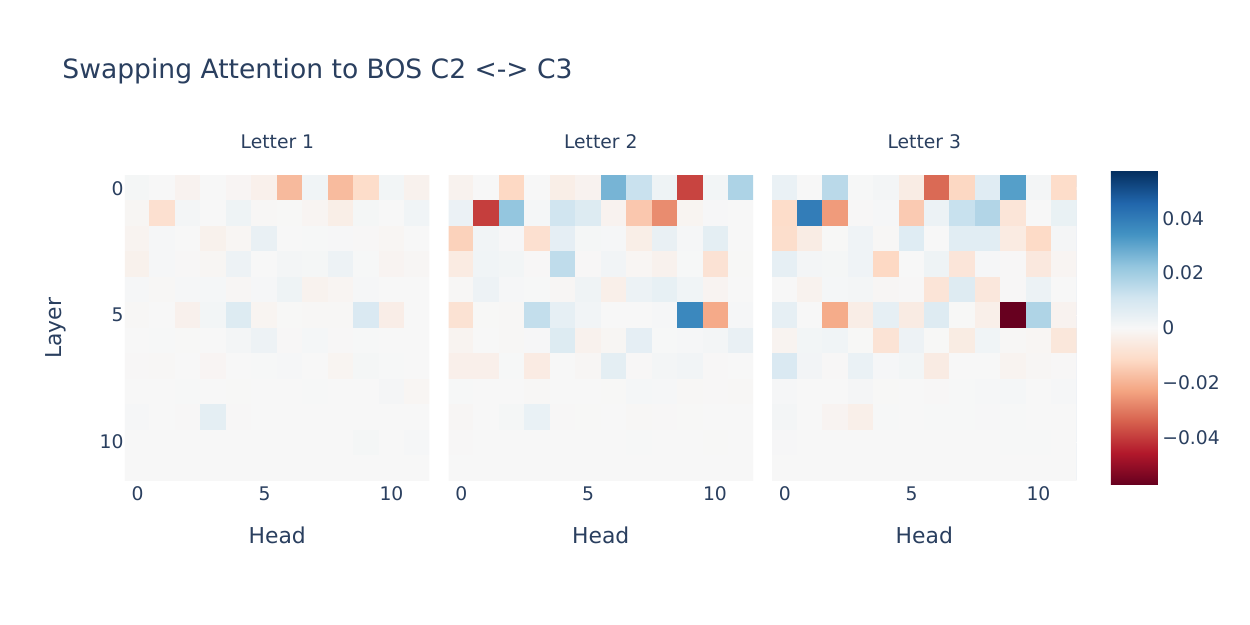}
    \caption{Change in logit difference obtained by swapping the attention paid to \texttt{BOS} from the \texttt{C2} and \texttt{C3} tokens for every head in the model.}
    \label{fig:swap_BOS_C2_C3_attn_patching}
\end{figure}

This section contains the remaining positional experiments (presented in Section 4.2) regarding the rest of possible swapping combinations and letter mover heads. Figs. \ref{fig:swap_BOS_C1_C2_attn_patching} and \ref{fig:swap_BOS_C2_C3_attn_patching} show the result of swapping the attention paid to \texttt{BOS} from the \texttt{C1} and \texttt{C2} tokens, and the \texttt{C2} and \texttt{C3} tokens respectively. 

\begin{figure}[htbp]
\centering
\begin{tabular}{cc}
  \includegraphics[width=0.45\linewidth]{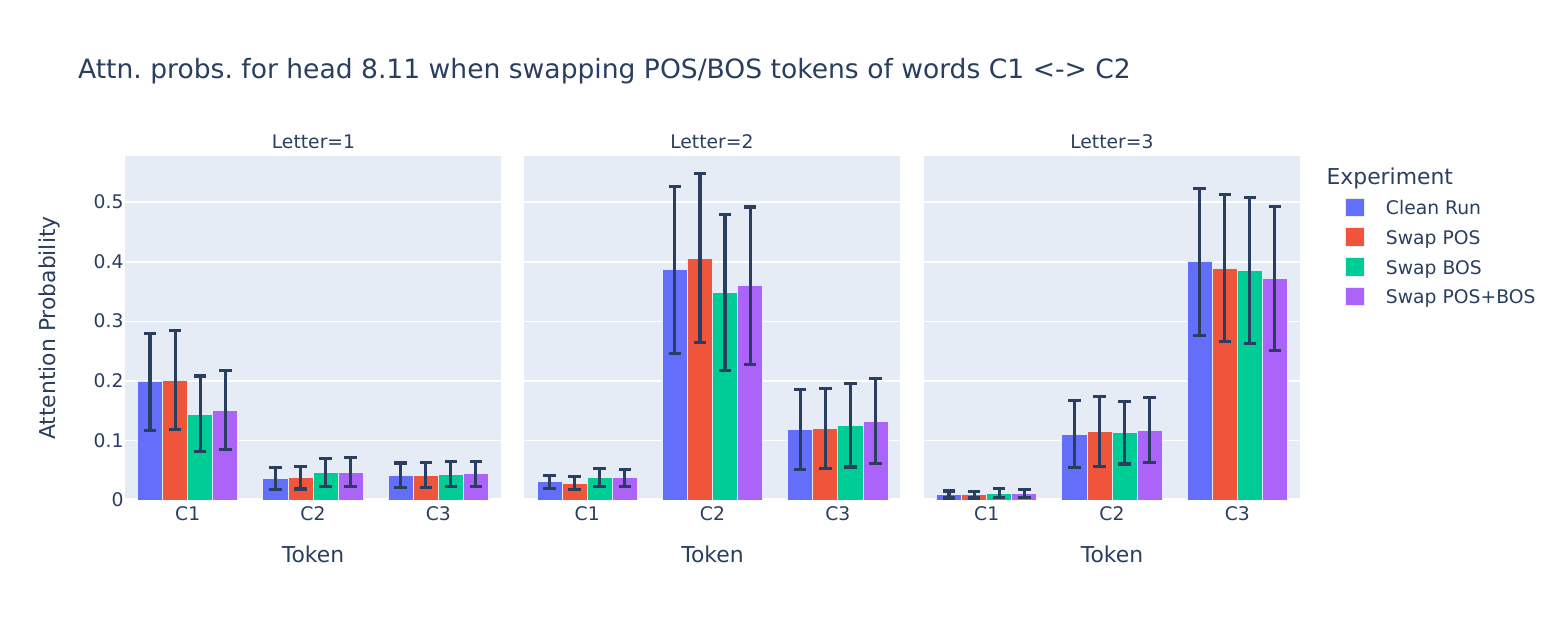} &   \includegraphics[width=0.45\linewidth]{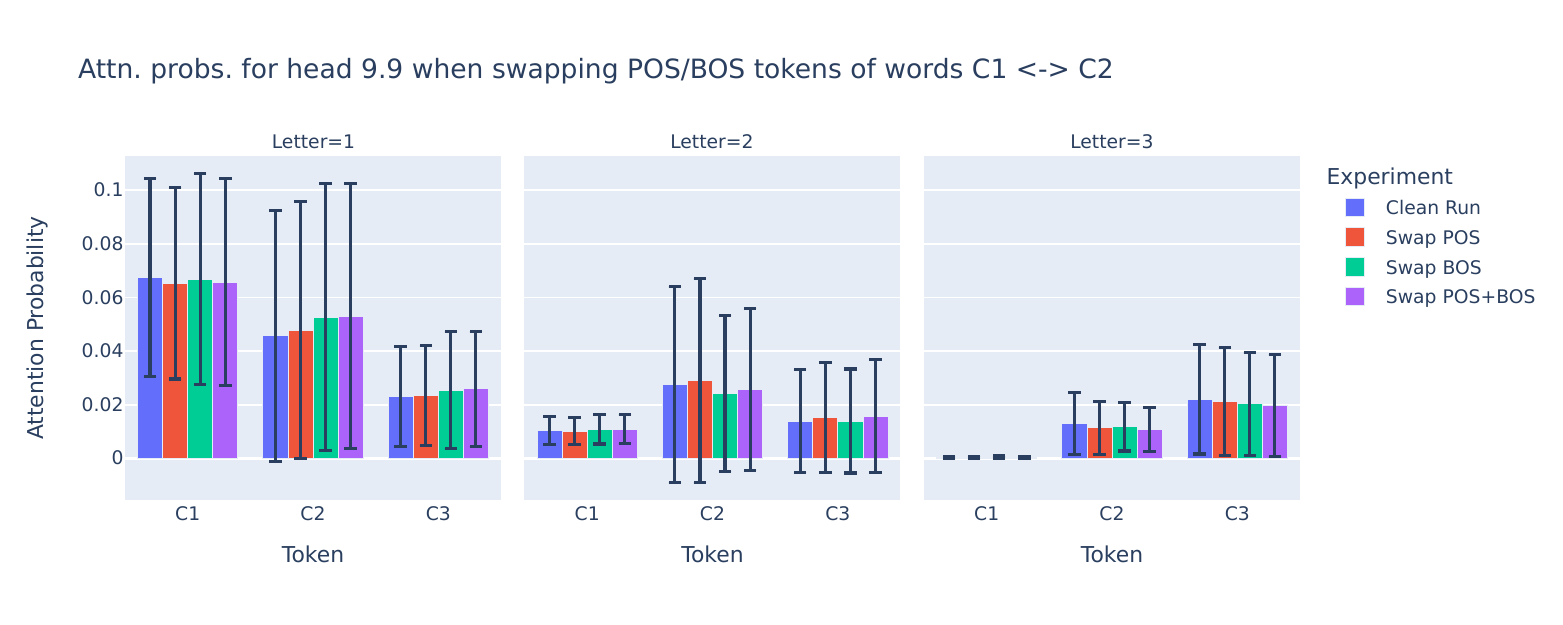} \\
  \includegraphics[width=0.45\linewidth]{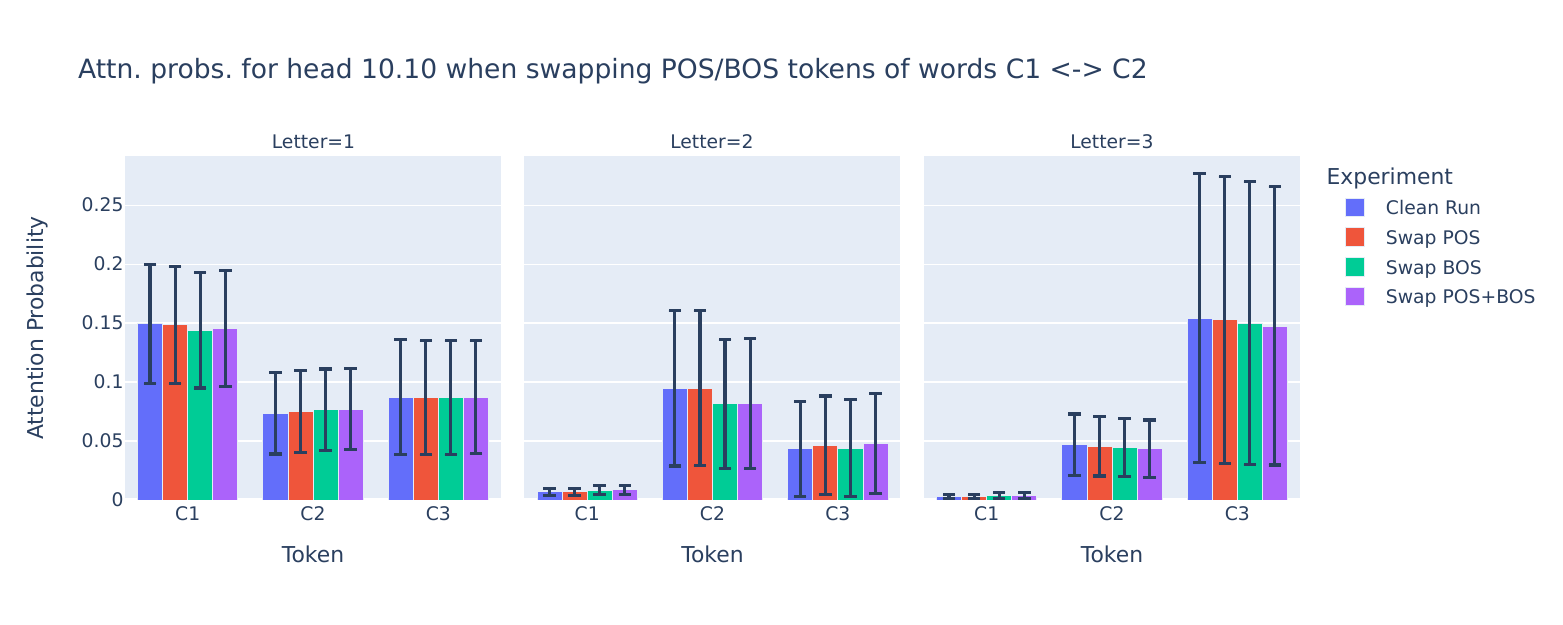} &   \includegraphics[width=0.45\linewidth]{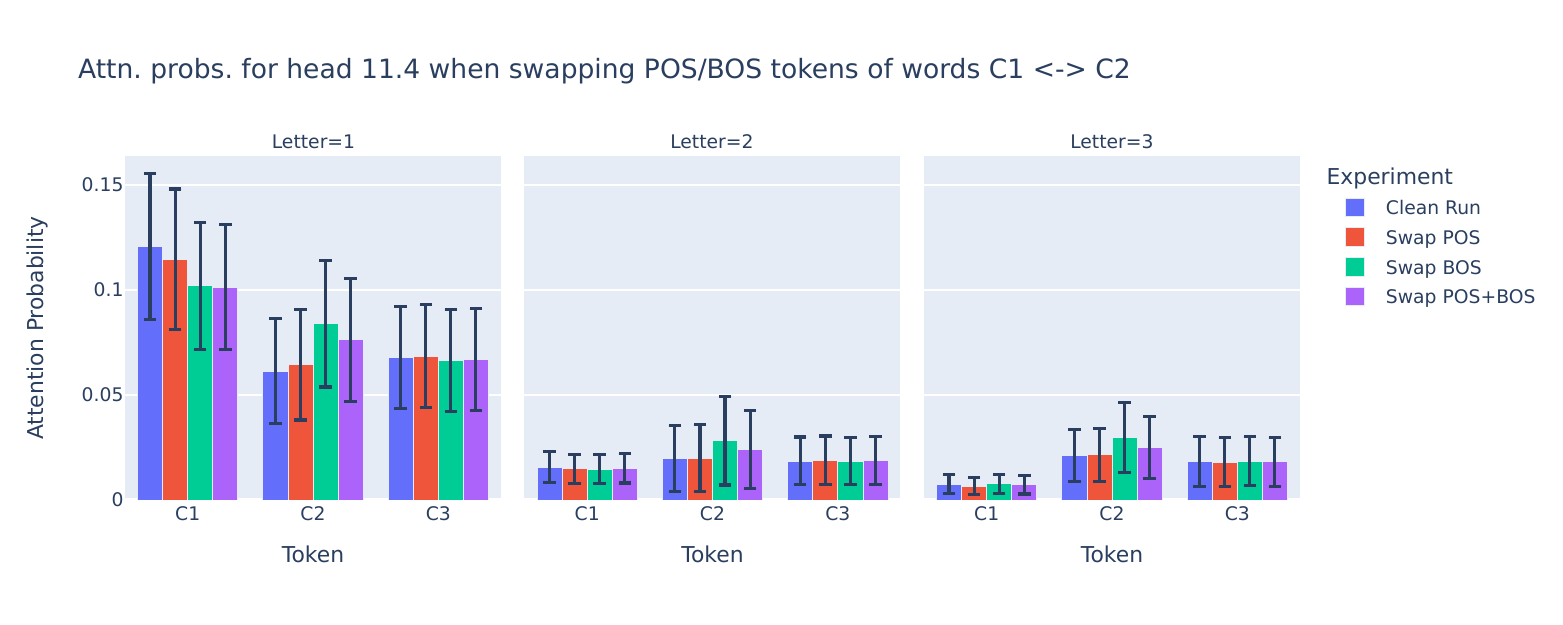} \\
\end{tabular}
\caption{Effect of swapping the positional embeddings and/or attention to \texttt{BOS} of \texttt{C1} and \texttt{C2} on the attention paid to the capital letter tokens for each letter mover head.}
\label{fig:swap_1_2}
\end{figure}

\begin{figure}[htbp]
\centering
\begin{tabular}{cc}
  \includegraphics[width=0.45\linewidth]{BOS_8_11_1_3.pdf} &   \includegraphics[width=0.45\linewidth]{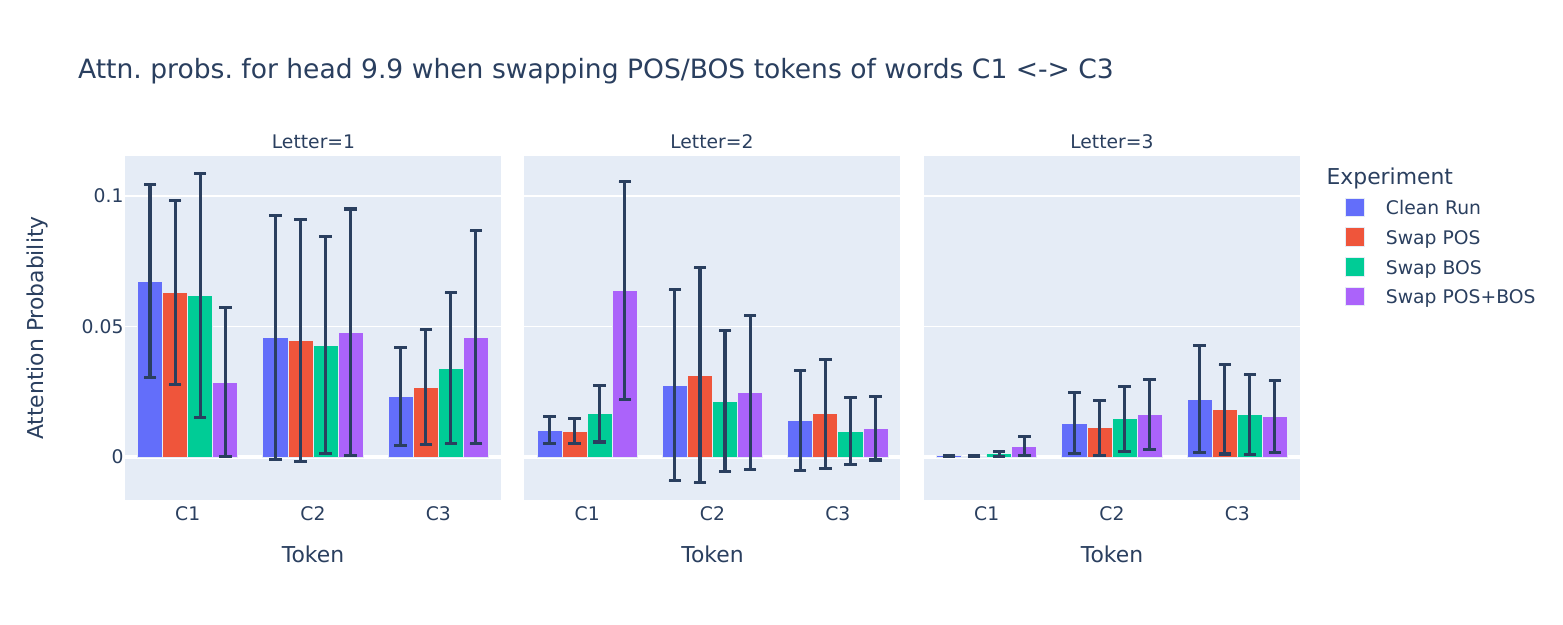} \\
  \includegraphics[width=0.45\linewidth]{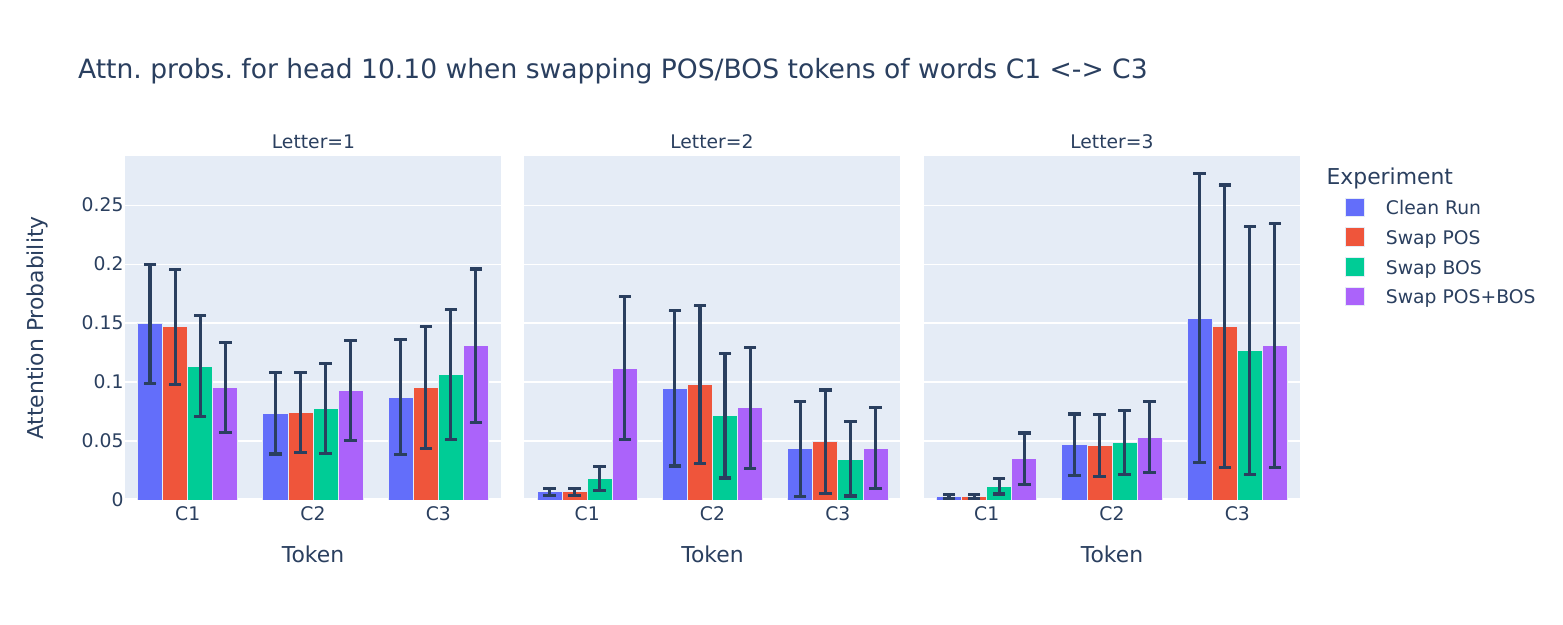} &   \includegraphics[width=0.45\linewidth]{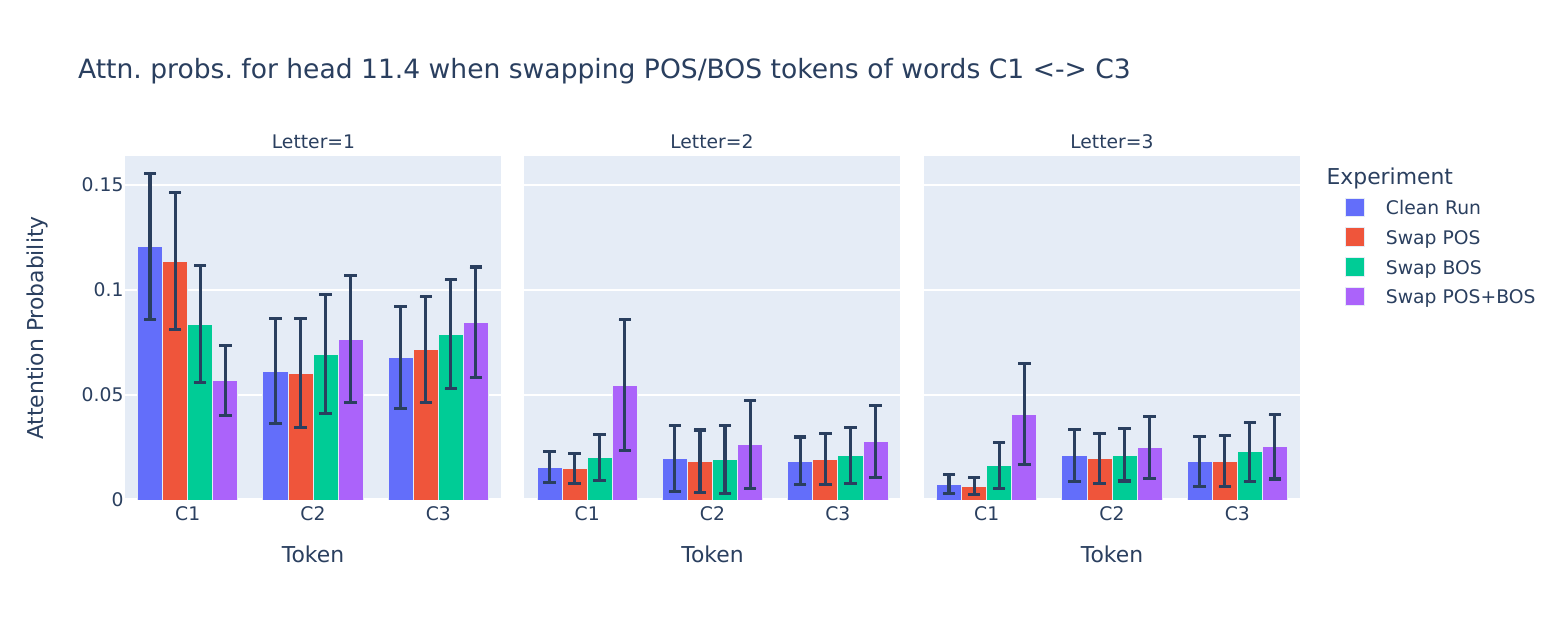} \\
\end{tabular}
\caption{Effect of swapping the positional embeddings and/or attention to \texttt{BOS} of \texttt{C1} and \texttt{C3} on the attention paid to the capital letter tokens for each letter mover head.}
\label{fig:swap_1_3}
\end{figure}

\begin{figure}[htbp]
\centering
\begin{tabular}{cc}
  \includegraphics[width=0.45\linewidth]{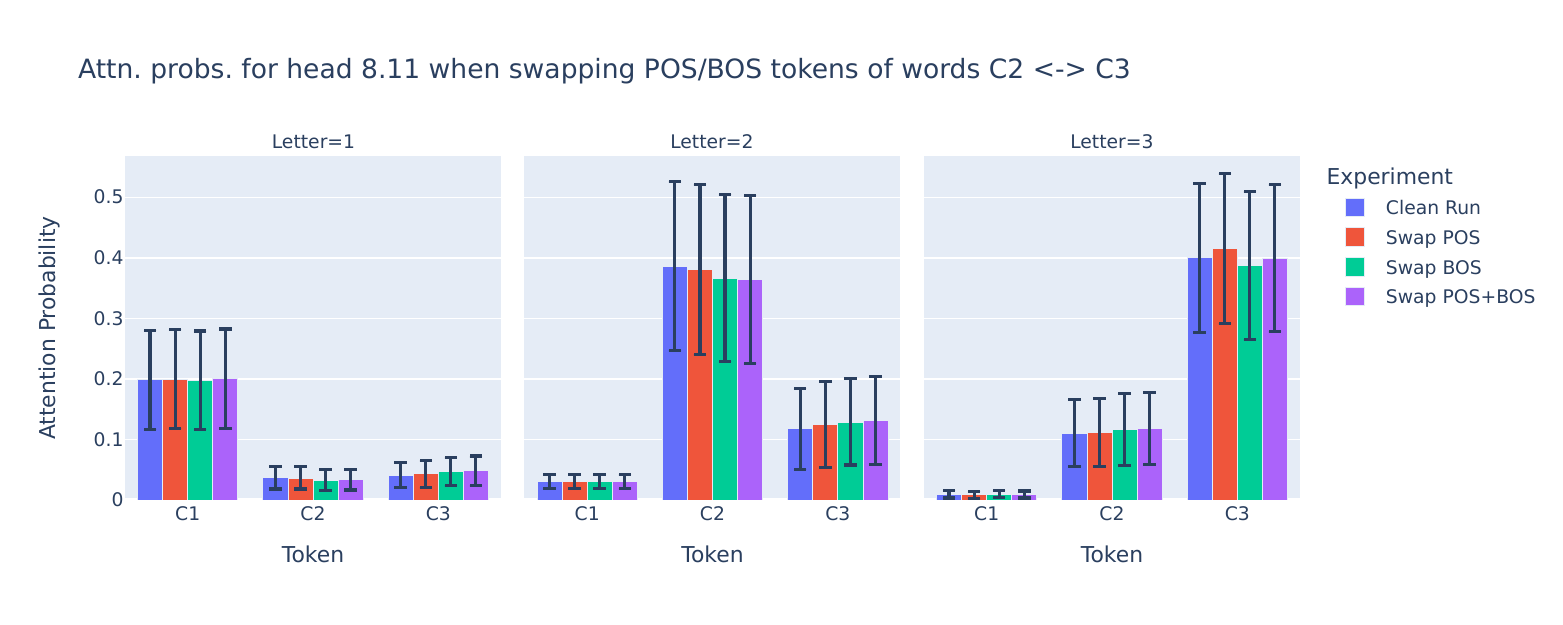} &   \includegraphics[width=0.45\linewidth]{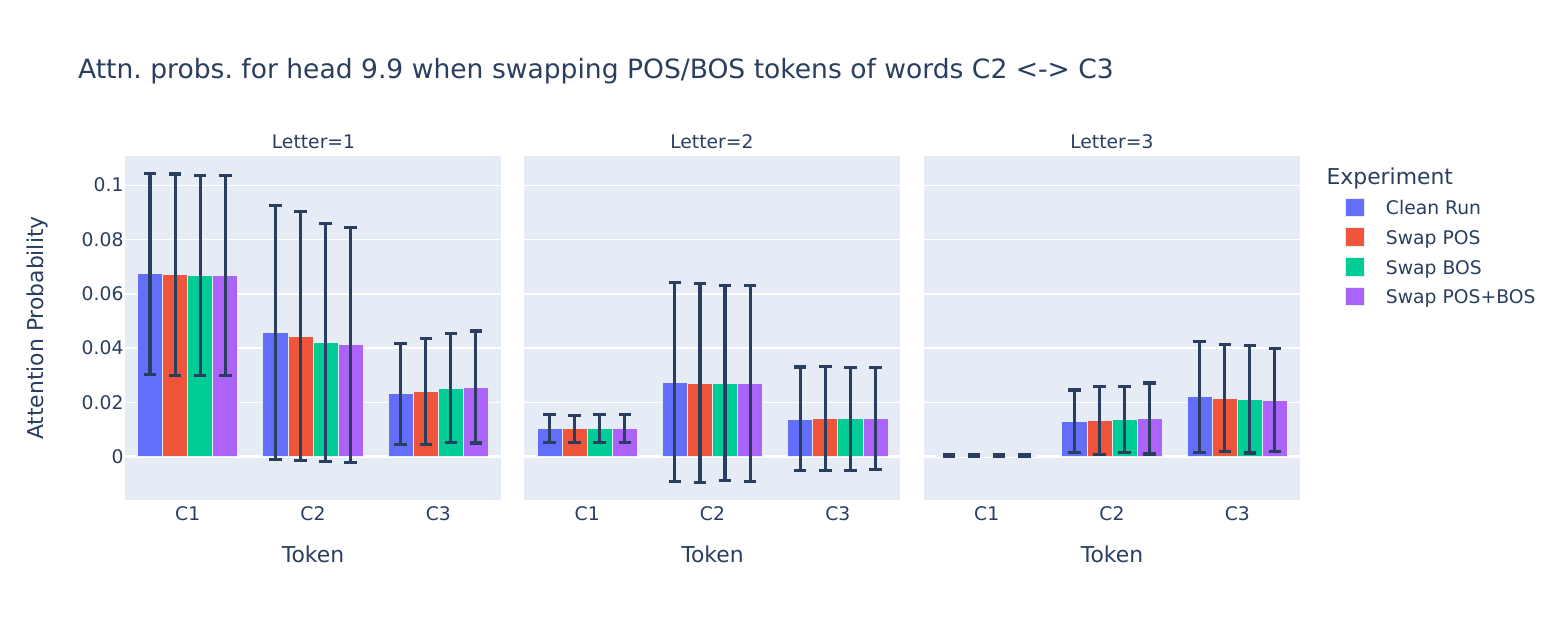} \\
  \includegraphics[width=0.45\linewidth]{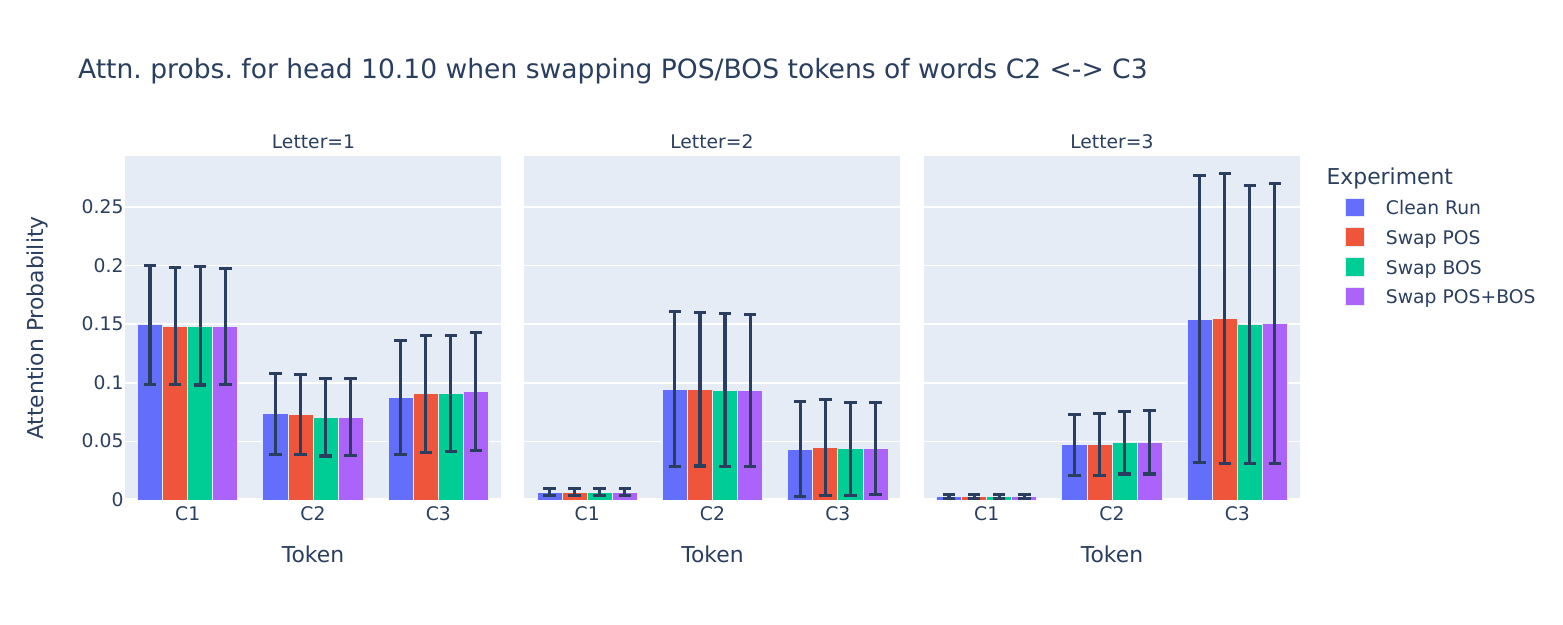} &   \includegraphics[width=0.45\linewidth]{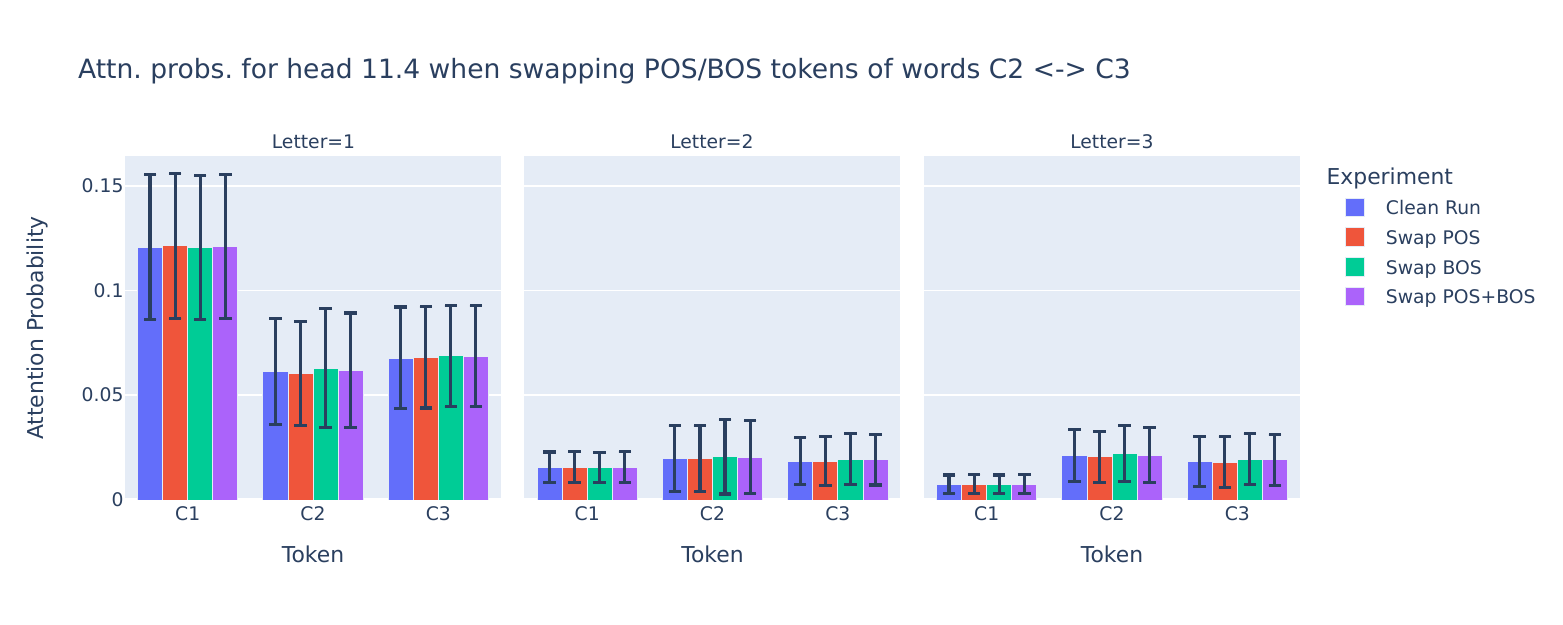} \\
\end{tabular}
\caption{Effect of swapping the positional embeddings and/or attention to \texttt{BOS} of \texttt{C2} and \texttt{C3} on the attention paid to the capital letter tokens for each letter mover head.}
\label{fig:swap_2_3}
\end{figure}

Figs. \ref{fig:swap_1_2}-\ref{fig:swap_2_3} show the difference in attention paid to the \texttt{Ci} tokens on the clean run, when swapping the positional embeddings, swapping the attention paid to the \texttt{BOS} token, and performing both swaps at the same time. This is visualized for every possible swap and letter mover head. As mentioned in the paper, it can be seen that the largest impact happens when performing the swapping operation on tokens \texttt{C1} and \texttt{C3}, specially on the first letter prediction, whereas the changes on the other swapping experiments are almost negligible.

\end{document}